%% file: main.tex
\documentclass[lettersize,journal,onecolumn]{IEEEtran}
\usepackage{amsmath, amsfonts, algorithmic, algorithm, array, graphicx, cite, url, xcolor,lipsum, booktabs, multirow, makecell, enumitem, xurl, cite, xspace, makecell}
\usepackage[caption=false,font=footnotesize]{subfig}
\usepackage[hidelinks]{hyperref}
\hypersetup{
    colorlinks=true,  
    urlcolor=blue,     
    linkcolor=red,    
    citecolor=green     
}

\newcommand{\FoneW}{\text{F1}\textsubscript{w}}
\newcommand{\mAP}{\text{mAP}}
\newcommand{\AP}[1]{\text{AP}\textsubscript{#1}}
\newcommand{\APc}{\AP{c}\xspace}
\newcommand{\PrecW}{\text{Prec}\textsubscript{w}}
\newcommand{\RecallW}{\text{Rec}\textsubscript{w}}

\newcommand{\SWIRBetter}{\text{SWIR}\textsuperscript{+}\xspace}
\newcommand{\RGBBetter}{\text{RGB}\textsuperscript{+}\xspace}

\begin{document}

\title{%
  {\normalsize\color{gray}\normalfont
   This work has been submitted to the IEEE for possible publication.\\[-16pt]
   Copyright may be transferred without notice, after which this version
   may no longer be accessible.}\\[8pt]
  Analysis of SWIR Imaging Detection Performance Under Adverse Environmental Conditions for Autonomous Driving Systems%
}

\author{
\vspace{5pt}
\begin{tabular}{@{}c@{\hspace{2.1em}}c@{\hspace{2.1em}}c@{\hspace{2.1em}}c@{\hspace{2.1em}}c@{}}
Rohan Mehra & 
Alexandre Riffard & 
Yannis Loumouamou & 
Mathieu Labussière \\
\end{tabular}
\\[2pt]  %
\normalsize
\begin{tabular}{@{}c@{}}
Universit\'{e} Clermont Auvergne, Clermont Auvergne INP, CNRS, Institut Pascal, F-63000 Clermont-Ferrand, France\\
\end{tabular}
\\[-2pt]  
\footnotesize
\begin{tabular}{@{}l@{}}
\texttt{romehraa@gmail.com} \hspace{3.9em}
\texttt{\{firstname\}.\{lastname\}@uca.fr} 
\end{tabular}
}

\maketitle

\input{sections/00_abstract}

\input{sections/01_intro}

\input{sections/02_methodology}

\input{sections/03_global_results}
\input{sections/04_sensor_dominance}
\input{sections/05_discussion}
\input{sections/06_conclusion}

\bibliographystyle{IEEEtran}
\bibliography{references.bib}

\end{document}

%% file: sections/00_abstract.tex
\begin{abstract}
 Short-wave infrared (SWIR) imaging has emerged as a promising modality for autonomous driving, yet its practical benefits over RGB remain poorly characterized across diverse conditions. This paper presents a systematic comparative study of paired RGB and SWIR object detection on the RASMD dataset, covering four weather conditions and two real-time detection architectures, with various fine-tunings evaluated against a unified ground truth. Overall, RGB demonstrates comparable or superior performance in most scenarios, while RF-DETR exhibits greater robustness across varying conditions. Beyond aggregate metrics, we propose a sensor-dominance mining framework that combines multi-model agreement with targeted manual inspection to identify scenarios where one sensing modality provides more reliable detections using largely unannotated paired data. This analysis reveals that SWIR offers clear advantages in four safety-critical situations, including windshield glare, water droplets on the windshield, low-contrast object visibility, and long-range vehicle detection. The findings suggest that SWIR should be viewed as a complementary modality that enhances perception in rare but challenging conditions. \textit{The datasets will be available upon request, and all code and trained model weights are publicly released at \url{https://github.com/comsee-research/swir-adverse-env-analysis}.}
\end{abstract}

%% file: sections/01_intro.tex
\begin{figure}[!h] 
    \centering
    \vspace*{-2mm}\subfloat[Sunny]{
        \includegraphics[width=0.3\columnwidth]{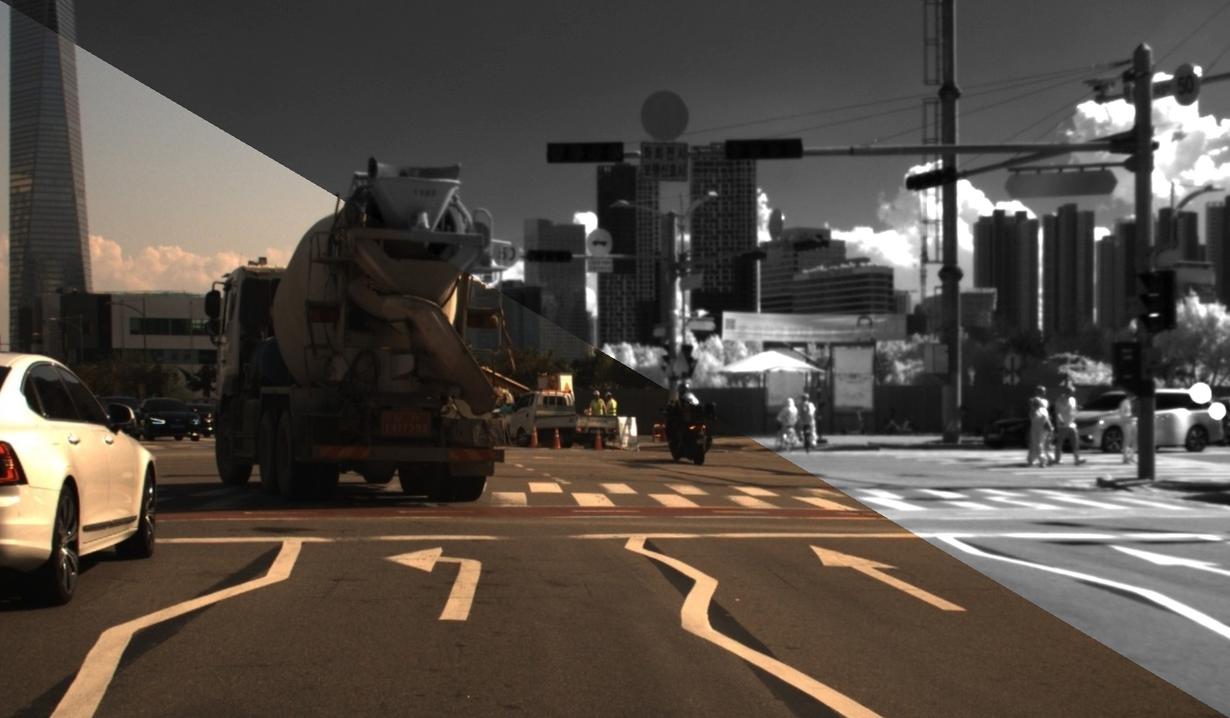}
        \label{fig:starred_sunny}
    }
    \subfloat[Cloudy]{
        \includegraphics[width=0.3\columnwidth]{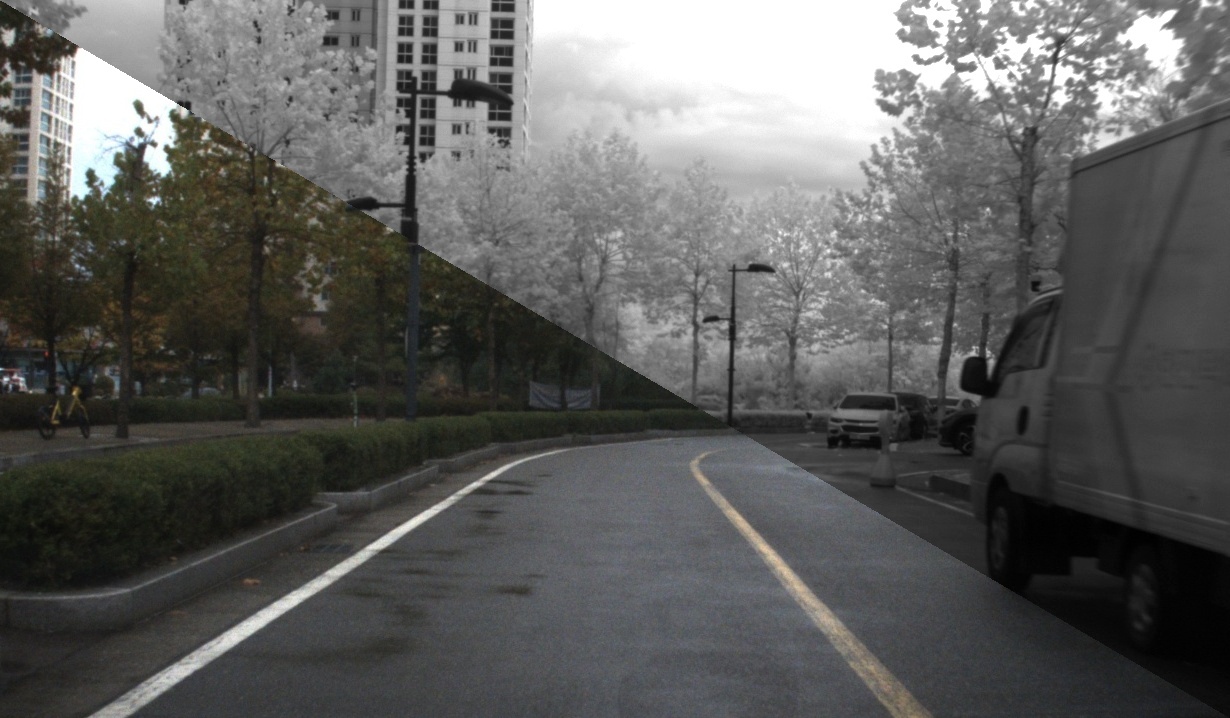} 
        \label{fig:starred_cloudy}
    }
    
    \vspace*{-0.2cm} 
    
    \subfloat[Rainy]{
        \includegraphics[width=0.3\columnwidth]{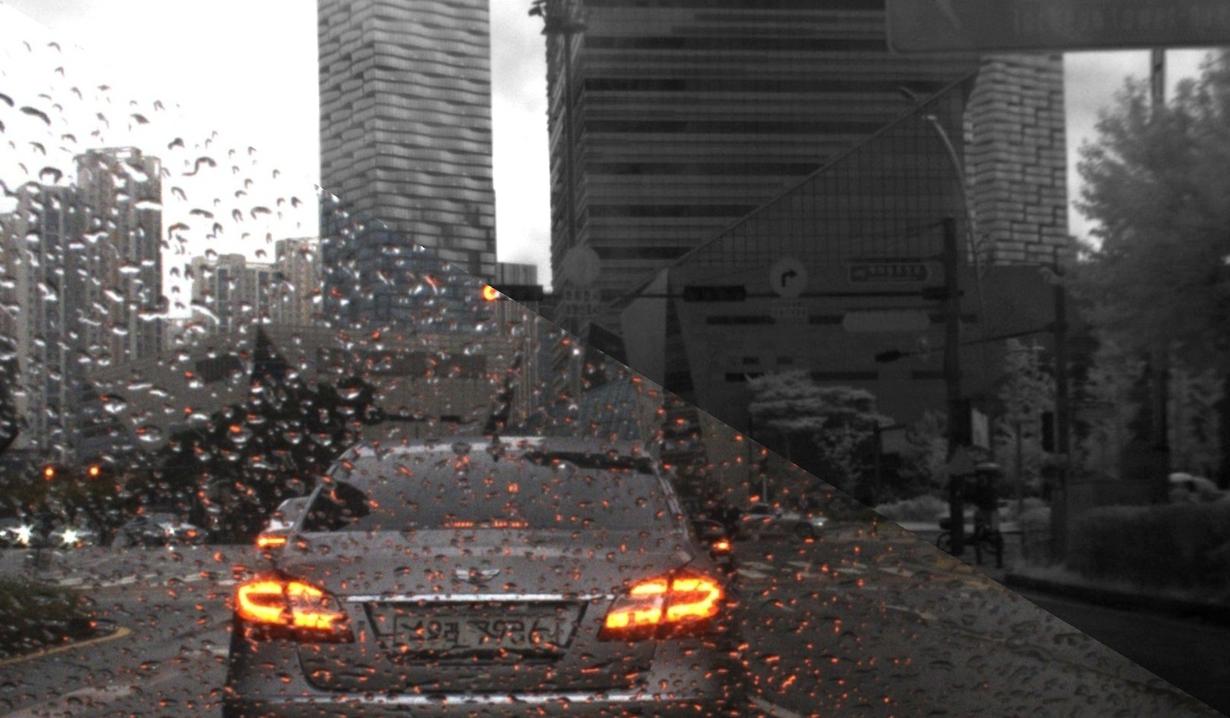} 
        \label{fig:starred_rainy}
    }
    \subfloat[Snowy]{
        \includegraphics[width=0.3\columnwidth]{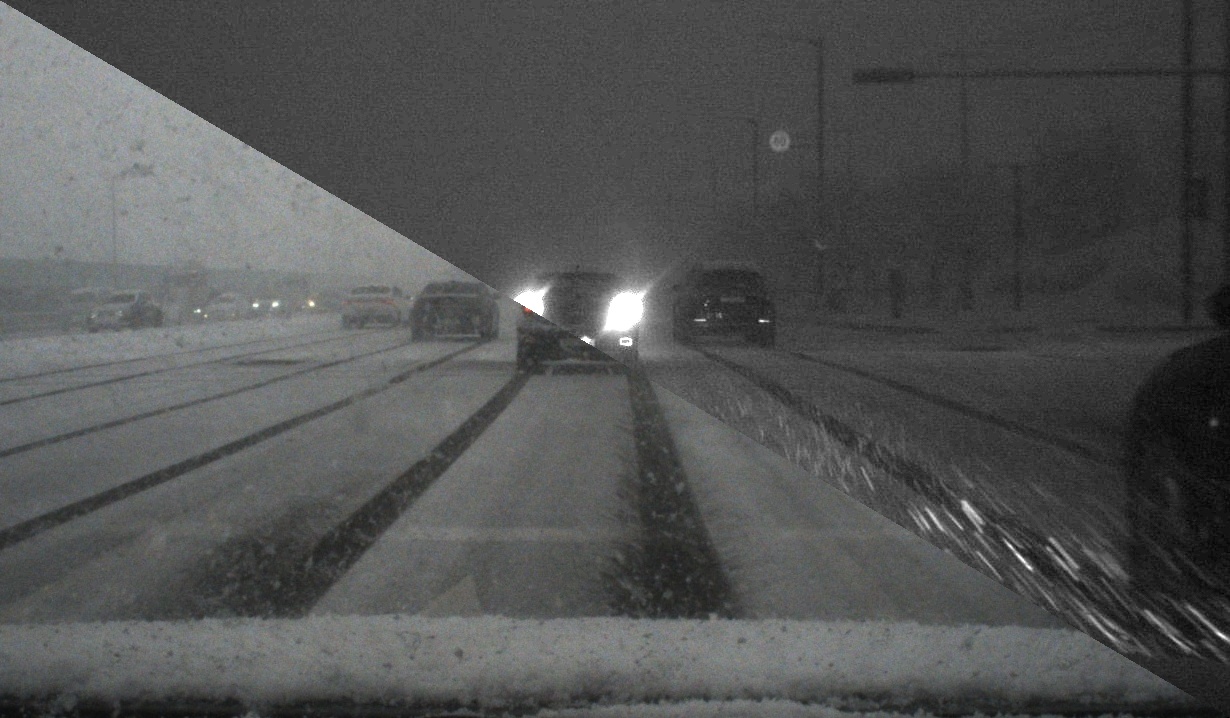}
        \label{fig:starred_snowy}
    }
    
    \caption{Paired RGB-SWIR images from the RASMD dataset \cite{JIN2026103872} across different weather conditions: (a) Sunny, (b) Cloudy, (c) Rainy, and (d) Snowy. Each image presents a diagonal split showing RGB (bottom-left) and SWIR (top-right).}
    \label{fig:starred_grid}
    \vspace*{-2mm}
\end{figure}

\section{Introduction}
\IEEEPARstart{A}{utonomous} driving relies heavily on robust environmental perception for safe navigation in dynamic road conditions. 
In current systems, this perception is predominantly achieved using visible-spectrum cameras, which suffer severe performance degradation in scenarios like low-light, fog, and adverse weather \cite{s20226532, 9157107, 10167647}. 
To close this gap, integration of visible with Radar \cite{zhou2022towards}, LiDAR \cite{li2020lidar}, Thermal (LWIR) \cite{7298706}, NIR \cite{articlenir}, or deploying computationally heavy algorithmic enhancements like image dehazing have been tried \cite{9561076}. 
However, these are accompanied by high costs, resolution trade-offs, and often fail in adverse conditions. 
Short-Wave Infrared (SWIR) \cite{10.1117/12.777776}, especially with the advent of CQD sensing technology \cite{cqdSID2021}, is becoming increasingly accessible as a viable alternative, providing distinct advantages compared to traditional sensors. 

Broader sensor surveys have positioned SWIR as a compelling candidate among emerging imaging modalities for autonomous vehicles. 
Its unique spectral properties and complementary role alongside LiDAR, radar, LWIR, and NIR have been highlighted in~\cite{li2023emergent} and~\cite{driggers2013good}. 
Empirical evidence indicates that SWIR maintains stronger robustness than visible-spectrum cameras under degraded visibility conditions such as fog, rain, and low illumination~\cite{10.1117/12.2519423,pinchon2018all,bessonov2025all, RPIAF, jobert2024swir}. 
In parallel, recent works have explored SWIR for detection and segmentation tasks. 
For instance, the authors of \cite{s22072562} used cross-spectral mapping to generate SWIR training data to significantly improve deep-learning object detection models. 
Additionally, in {preliminary work} \cite{11375753}, we showed that SWIR imagery enables pre-trained models to outperform standard cameras for object detection and segmentation in dense fog, using a small dataset acquired in controlled weather conditions.
Highlighting its long-range capabilities, in \cite{bertozzi2013pedestrian} they show that SWIR wavelengths penetrate haze to provide clearer pedestrian detection at a distance.

\begin{figure*}[!t]
    \centering
    \includegraphics[width=0.95\textwidth]{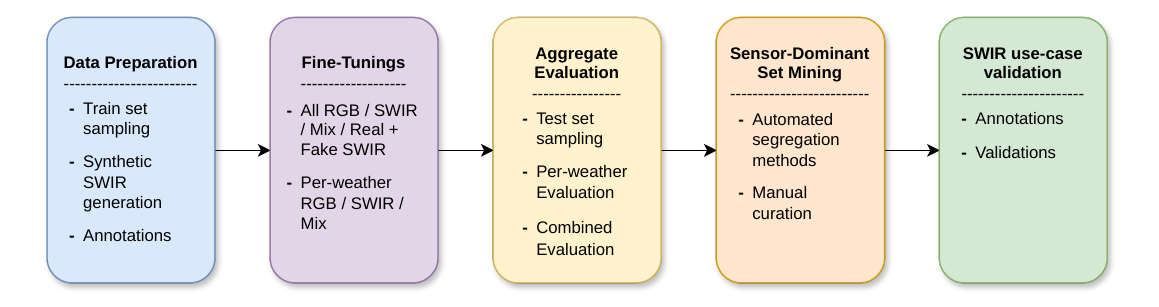}\vspace*{-4mm}
    \caption{High-level overview of the experimental pipeline. The block diagram illustrates the progression of tasks undertaken in this study, which are discussed in detail in the subsequent sections.}
    \label{fig:flow_diagram}\vspace*{-2mm}
\end{figure*}

Despite these advances, existing studies largely focus only on a specific condition or all-combined evaluations, primarily due to limited data availability. 
The introduction of the RASMD dataset~\cite{JIN2026103872}, the first large-scale paired RGB-SWIR set acquired in various weather conditions, as illustrated in \autoref{fig:starred_grid}, provided a benchmark across various tasks to analyze general SWIR capacities. 
Subsequently, an empirical study leveraged this data to demonstrate SWIR's optical advantages specifically under rainy conditions~\cite{YEO2026119654}. 
Building on this, the availability of the dataset enables a more systematic investigation of cross-modal detection performance and motivates the need to identify exact cases where SWIR provides practical benefits over RGB.\\

\noindent\textbf{Contributions.} This work makes the following contributions: \textbf{(i)} a comprehensive detection evaluation comparing paired RGB and SWIR images across four weather conditions, two real-time architectures, and five fine-tuning strategies against a unified ground truth; 
\textbf{(ii)} a \textit{sensor-dominance mining} framework combining multi-model consensus with human curation to segregate, at scale, scenarios where a particular modality yields superior detections from unannotated paired data; 
and \textbf{(iii)} identification and validation of four safety-critical SWIR-advantageous scenarios: glare, windshield water droplets, hidden objects under low-contrast, and distant vehicles.

The two annotated detection subsets with unified-modality ground truth: RASMD-IP, a weather-partitioned split for model training and evaluation, and RASMD-HiLS, a curated subset of SWIR-beneficial scenes categorized by identified conditions, are both available upon request. All code and trained model weights are publicly released at \url{https://github.com/comsee-research/swir-adverse-env-analysis}.

%% file: sections/02_methodology.tex
\section{Methodology}
\label{sec:methodology}

In this study, we evaluate different fine-tuning strategies for CNN- and Transformer-based object detection models on paired RGB and SWIR images captured under dynamic scenarios across four weather conditions. 
An overview of the complete pipeline is shown in \autoref{fig:flow_diagram}. This section outlines the dataset construction, fine-tuning strategies, and evaluation protocol used for model training and testing.

\subsection{Datasets}
\label{subsec:dataset}

We utilized as base the aligned subset images from the RASMD dataset \cite{JIN2026103872}, containing approximately 100,000 paired RGB-SWIR driving images divided into sunny, cloudy, rainy, and snowy conditions. 
Although the original RASMD dataset provides an additional set of annotated images for the detection task, these are not utilized in our study due to certain limitations. In particular, the frequently lacking consistent spatial alignment and poor pairing of images between RGB and SWIR modalities can lead to object localization discrepancies. Moreover, the absence of explicit weather-wise division and limited scene diversity further limit its suitability for our evaluation. 
A newer version of the detection subset (RASMD-detect-v2) addresses some of these issues; however, it was not available at the time of this study. 
Therefore, we built our own detection datasets from the original version.\\

\noindent\textbf{RASMD-IP.}~To evaluate general detection performance, we uniformly sample across all urban and suburban sequences in \emph{each weather condition}, constructing a training set of 1,713 paired images, ensuring that the model is exposed to the full range of available scenes. This set is subsequently split into training and validation subsets. The test set, consisting of 800 paired images (i.e., 200 per weather condition), is sampled following the procedure described in \autoref{subsec:eval_protocol}, to ensure sufficient supports of each class for significant evaluation.
\autoref{tab:class_support} summarizes the class-wise distribution of samples in this dataset. \\

\begin{table}[t]
    \centering
    \renewcommand{\arraystretch}{1.2} 
    \setlength{\tabcolsep}{5pt} 
    
    \caption{Class-wise distributions in the RASMD-IP dataset. Training set cover all weather and scenes uniformly, whereas the test set prioritizes per-class support for significant evaluations.}
    \label{tab:class_support}
    
    \begin{tabular}{l r r r}
        \toprule
        \textbf{Class}
        & \makecell[r]{\textbf{Train} \\ {\scriptsize \textit{(1713 images)}}} 
        & \makecell[r]{\textbf{Test} \\ {\scriptsize \textit{(800 images)}}} 
        & \makecell[r]{\textbf{Total} \\ {\scriptsize \textit{(2513 images)}}} \\
        \midrule\midrule
        Car        & 7,531 & 5,092 & 12,623 \\
        Truck      & 1,845 & 897   & 2,742  \\
        Person     & 480   & 828   & 1,308  \\
        Bus        & 421   & 566   & 987    \\
        Motorcycle & 87    & 228   & 315    \\
        Bicycle    & 60    & 217   & 277    \\
        \midrule
        \textbf{Total (\# instances)} & \textbf{10,424} & \textbf{7,828} & \textbf{18,252} \\
        \bottomrule
    \end{tabular}
\end{table}

\noindent \textbf{RASMD-HiLS.}~To analyze scenarios where SWIR provides advantages over RGB, we introduce the {RASMD-HiLS} dataset. As detailed in \autoref{sec:hils}, a multi-model consensus framework followed by human curation is used to isolate 532 RGB–SWIR pairs exhibiting clear SWIR benefits. These samples fall into four recurring conditions: \textbf{(1)} windshield glare, \textbf{(2)} water droplets on the windshield, \textbf{(3)} low-contrast object visibility, and \textbf{(4)} distant vehicles. Each condition is further validated in \autoref{sec:swir-plus-valid}\\

\noindent\textbf{Annotations.} Images from both datasets were manually annotated using a unified ground truth, where an object is labeled if it is visible in either the RGB or SWIR image. Note that in RASMD annotated set \cite{JIN2026103872}, annotations are provided in modality-specific manner. 
Additionally, the SWIR images were preprocessed by clipping intensity values at $99\%$ confidence interval to suppress outliers and stretching the histogram to enhance contrast \cite{11375753}. \textit{Our annotated datasets will be made available upon request.}

\subsection{Synthetic SWIR Generation}
\label{subsec:syn-swir}

In SWIR imaging research, the limited availability of labeled datasets poses a significant challenge, which has led to approaches for generating synthetic images from RGB data. 
{An other work \cite{YannisRFIAP2026} shows promising use of synthetic SWIR for fine-tuning detection models.}
In this work, we also investigate the effectiveness of augmenting synthetic SWIR data alongside real SWIR data for fine-tuning detection models. 
Building on the benchmarks established in \cite{JIN2026103872}, we employ Pix2PixHD \cite{wang2018pix2pixHD} to translate 20\% of the RGB training images from the RASMD-IP dataset into the SWIR domain. 
Combined with real SWIR images, we compare the resulting performance to models fine-tuned purely on real SWIR data (see \autoref{sec:global_res}).

\subsection{Object detection Fine-Tuning Strategies}

In this study, we explore two real-time deep learning object detection architectures: 1) \textit{YOLOv8x} \cite{yolov8_ultralytics}, a CNN-based single-stage detector, and 2) \textit{RF-DETR} \cite{robinson2026rfdetrneuralarchitecturesearch}, built on a ViT backbone. Both models, pre-trained on the COCO dataset \cite{cocodataset}, are fine-tuned on the following configurations of our RASMD-IP training data:

\begin{itemize}%
\item \textbf{RGB-only images}, from all weather conditions, serving as the baseline aligned with the pre-training domain.
\item \textbf{SWIR-only images}, from all weather conditions, to evaluate standalone SWIR performance under domain shift.
\item \textbf{Synthetic SWIR (20\%), with real SWIR (80\%) images}, across all weather conditions, to explore the direction of expanding scene diversity in SWIR training using translated RGB data.
\item \textbf{Mixed RGB-SWIR images}, in equal proportion across all weather conditions, to study cross-modal learning within a single unified model.
\item \textbf{Per-weather RGB / SWIR / Mixed images}, where independent models are trained for each weather condition and modality. This is performed to examine condition-specific specialization versus generalization.
\end{itemize}
In the rest of the paper, the data splits will be respectively referred to as RGB, SWIR, SWIR-Syn, Mix, and in the format \texttt{[Weather]-[Modality]} for weather-specific cases. The corresponding fine-tuned models use the following nomenclature \texttt{[Model]-[Data]}, for instance, YOLOv8x on mixed data is YOLO-Mix. 

\subsection{Evaluation Protocol}
\label{subsec:eval_protocol}

The overall dataset is inherently subject to high class imbalance, which further increases when split by weather conditions (e.g., presence of motorcyclists is rare compared to that of cars in rainy weather). To ensure sufficient class support for maintaining statistical significance across analyses, we use a rarity-weighted sampling strategy to construct the test set (800 images, 200 per weather condition). Pseudo-labels were generated for all the 100,000 images in a single modality using our YOLO-Mix fine-tunned model which was used to guide the selection. Each image was assigned a score $S_i$ based on the inverse frequency of the detected classes, defined by

\begin{equation}
S_i = \sum_{c \in C} \frac{N_{i,c}}{G_c}\text{~,}
\end{equation}

\noindent where $C$ denotes the set of target classes, $N_{i,c}$ is the number of instances of class $c$ in image $i$, and $G_c$ is the total number of instances of class $c$ in the dataset.

To avoid near-identical frames, images within $\pm 10$ frames from the same sequence have not been selected together. Images were first selected in descending order of their scores, and any remaining slots were filled with randomly chosen images while maintaining the temporal constraint. \autoref{tab:class_support} shows the class-wise supports in the test set. Note that while most classes have sufficient support, some rare classes under specific weather conditions (e.g., bicyclists in snowy weather) remain underrepresented, limiting the reliability of their individual low-support class results.

\noindent\textbf{Metrics.} For evaluation, we compute Weighted Precision (\PrecW), Recall (\RecallW), and F1-score (\FoneW) deterministically on the RASMD-IP test images to account for the remaining class imbalance. Moreover, per-class average precision (\APc) and mean average precision (\mAP), has been calculated in COCO style using stratified bootstrapping \cite{e89fac9c-03d7-3e22-aa30-08f5596f8fce, NEURIPS2023_a0644215} with 1,000 iterations for stable analysis. Mean \APc is reported per class, while \mAP~is reported as mean $\pm$ standard deviation across weather conditions. All models are evaluated on their matching trained modality. Weather-specific models are evaluated on their matching weather subset only, while all-weather models are evaluated on both individual weather subsets and the combined test set.  Mix trained models are evaluated across both RGB and SWIR images.

%% file: sections/03_global_results.tex
\section{Global Results}
\label{sec:global_res}

This section presents the quantitative evaluation of RGB and SWIR sensor across all fine-tuning strategies, and has been organised as follows: combined analysis with aggregate metrics over all weather condition, and per-weather analysis comparing both modality across sunny, cloudy, rainy, and snowy conditions.

\begin{table*}[!b]
\centering
\caption{Quantitative evaluation across combined weather conditions for YOLOv8 and RF-DETR on paired RGB and SWIR data. Modality-specific fine-tuning improves performance over pre-trained models, while mixed fine-tuning gives competitive cross-modality performance.
\textbf{Bold} indicates the best, and \underline{underline} the the second best per camera.}
\setlength{\tabcolsep}{3.5pt}
\begin{tabular}{lll@{\hspace{1.5em}}ccc@{\hspace{1.5em}}ccccccc} 
 \toprule 
 \textbf{Camera} & \textbf{Model} & \textbf{Fine-Tuning} & \textbf{F1}\textsubscript{w} & 
 \textbf{Rec}\textsubscript{w} & \textbf{Prec}\textsubscript{w} & 
 \textbf{AP}\textsubscript{person} & 
 \textbf{AP}\textsubscript{car} & 
 \textbf{AP}\textsubscript{truck} & 
 \textbf{AP}\textsubscript{bus} & 
 \textbf{AP}\textsubscript{bicy.} & 
 \textbf{AP}\textsubscript{moto.} &
 \textbf{mAP} \\ 
 \midrule\midrule 
 
 \multirow{6}{*}{\makecell[l]{RGB}} & 
 \multirow{3}{*}{YOLOv8} & 
 PT & 0.6994 & 0.5841 & 0.8813 & \underline{0.3711} & 0.4793 & 0.3415 & 0.2764 & 0.2383 & 0.3014 & 0.3347 $\pm$ 0.0084 \\
 & & RGB & 0.7956 & 0.7357 & 0.8701 & 0.2945 & 0.5280 & 0.5329 & 0.4570 & 0.2401 & \underline{0.3247} & 0.3962 $\pm$ 0.0072 \\
 & & Mix & 0.7892 & 0.7214 & 0.8752 & 0.2505 & 0.5286 & 0.5131 & 0.4194 & 0.2149 & 0.2825 & 0.3681 $\pm$ 0.0072 \\ 
 
 \cmidrule(lr){2-13} 
 & \multirow{3}{*}{RF-DETR} & 
 PT & 0.7770 & 0.6905 & \textbf{0.8994} & \textbf{0.3973} & \underline{0.5556} & 0.4521 & 0.3600 & 0.2711 & \textbf{0.4340} & 0.4116 $\pm$ 0.0084 \\
 & & RGB & \textbf{0.8176} & \textbf{0.7634} & 0.8842 & 0.3518 & \textbf{0.5694} & \textbf{0.5680} & \textbf{0.4952} & \textbf{0.3442} & 0.2960 & \textbf{0.4374 $\pm$ 0.0075} \\
 & & Mix & \underline{0.8136} & \underline{0.7494} & \underline{0.8951} & 0.3113 & 0.5425 & \underline{0.5547} & \underline{0.4600} & \underline{0.2822} & 0.3122 & 0.4105 $\pm$ 0.0074 \\ 
 \cmidrule(lr){1-13} 
 \multirow{8}{*}{\makecell[l]{SWIR}} & 
 \multirow{4}{*}{YOLO} & 
 PT & 0.6189 & 0.4926 & 0.8446 & 0.2144 & 0.3416 & 0.2534 & 0.2191 & 0.1441 & 0.1456 & 0.2197 $\pm$ 0.0066 \\ 
 & & SWIR & 0.7580 & 0.6884 & 0.8475 & 0.1764 & 0.4432 & 0.4794 & 0.4135 & 0.1968 & 0.2584 & 0.3279 $\pm$ 0.0070 \\
 & & SWIR-Syn & 0.7485 & 0.6803 & 0.8355 & 0.1642 & 0.4341 & 0.4646 & 0.4043 & 0.2059 & 0.2416 & 0.3191 $\pm$ 0.0069 \\ 
 & & Mix & 0.7398 & 0.6984 & 0.8003 & 0.1443 & 0.4357 & 0.4704 & 0.4339 & 0.2123 & 0.2564 & 0.3255 $\pm$ 0.0068 \\ 
 \cmidrule(lr){2-13} 
 & \multirow{4}{*}{RF-DETR} & 
 PT & 0.6947 & 0.5876 & 0.8607 & \textbf{0.2304} & 0.3919 & 0.3384 & 0.2644 & 0.1547 & 0.2527 & 0.2721 $\pm$ 0.0072 \\ 
 & & SWIR & \textbf{0.7965} & \textbf{0.7320} & 0.8769 & \underline{0.2180} & \textbf{0.4704} & \underline{0.4878} & \textbf{0.4779} & \textbf{0.2513} & \underline{0.2853} & \textbf{0.3651 $\pm$ 0.0071} \\
 & & SWIR-Syn & \underline{0.7892} & \underline{0.7149} & \textbf{0.8852} & 0.2041 & \underline{0.4611} & 0.4868 & 0.4501 & \underline{0.2486} & \textbf{0.2879} & 0.3564 $\pm$ 0.0068 \\
 & & Mix & 0.7866 & 0.7137 & \underline{0.8812} & 0.1882 & 0.4506 & \textbf{0.4946} & \underline{0.4517} & 0.1941 & 0.2800 & 0.3432 $\pm$ 0.0066 \\ 
 \bottomrule\\
 \multicolumn{13}{l}{\textit{Abbreviations}: \textbf{PT} = Pre-trained; \textbf{Mix} = Mixed (50\%-50\%) RGB-SWIR Images; \textbf{SWIR-Syn} = Synthetic (20\%) + Real (80\%) SWIR Images.} 
 \end{tabular}
\label{tab:quantitative_evaluation_merged}
\end{table*}

\begin{figure*}[!t]
    \centering
    \vspace*{-4mm}\subfloat[Precision (\PrecW)]{
        \includegraphics[width=0.48\linewidth]{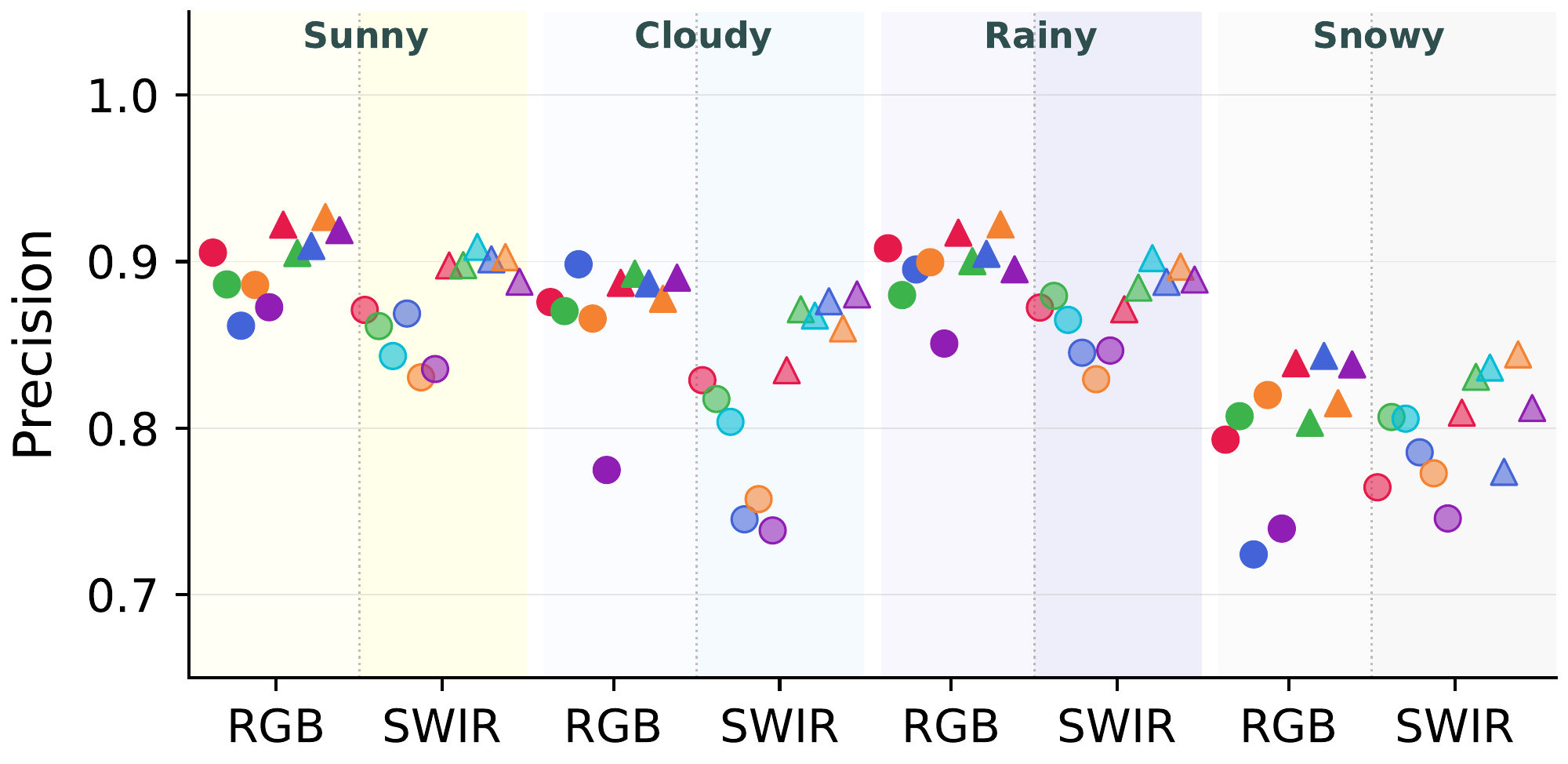}
        \label{fig:precision}
    }
    \hfil 
    \subfloat[Recall (\RecallW)]{
        \includegraphics[width=0.48\linewidth]{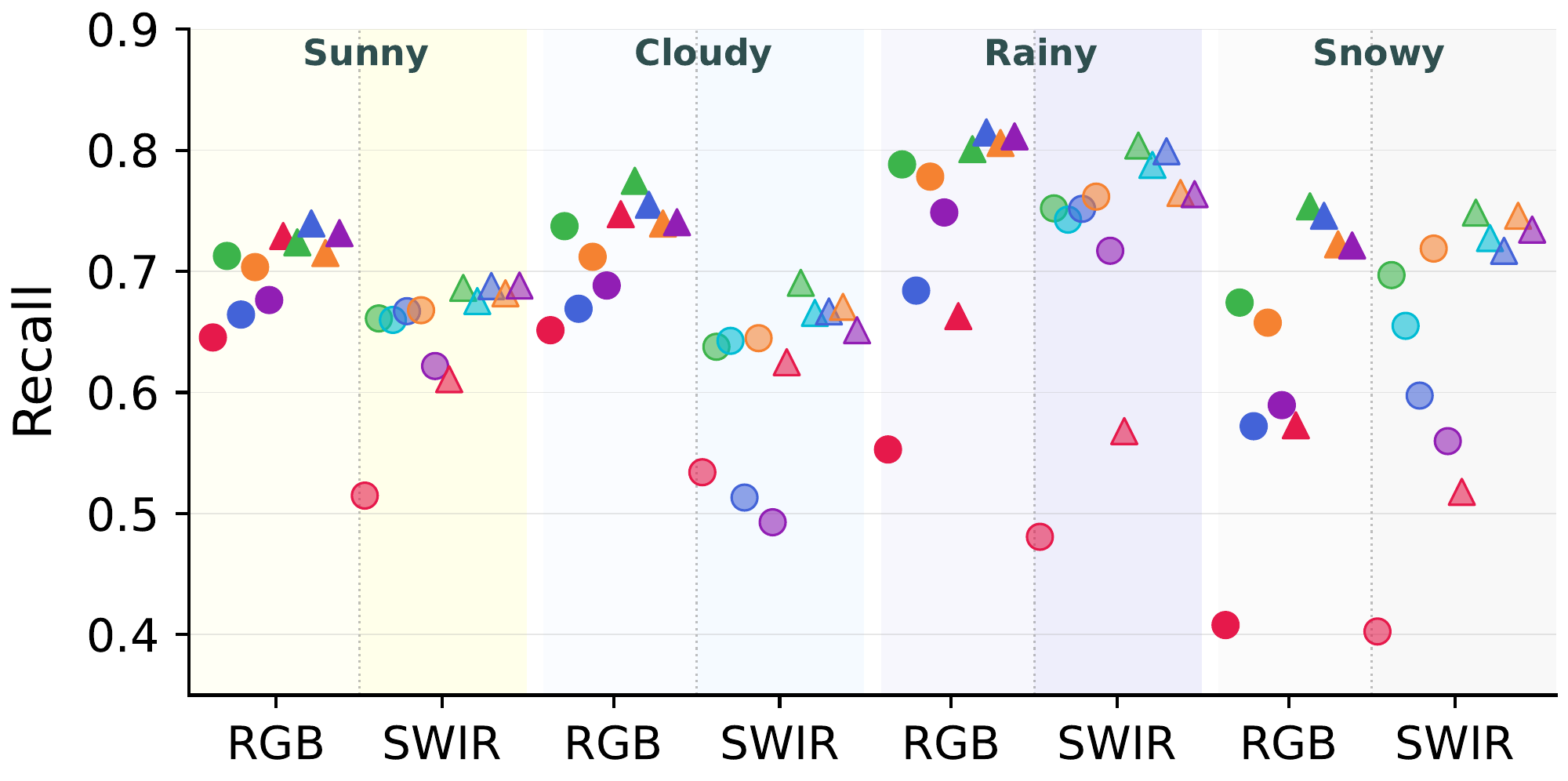}
        \label{fig:recall}
    }
    \vspace*{-2mm}
    \subfloat[F-1 Score (\FoneW)]{
        \includegraphics[width=0.48\linewidth]{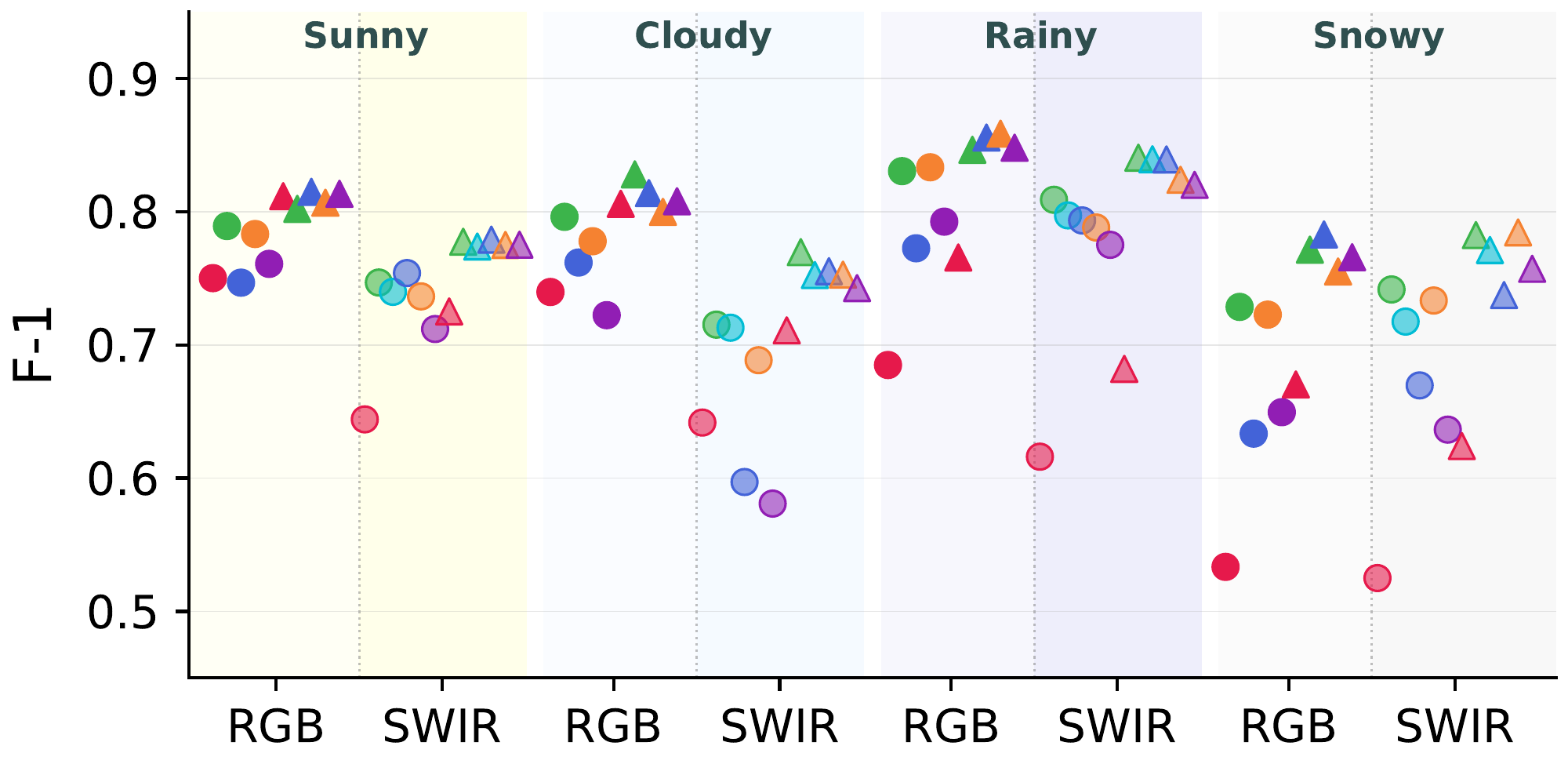}
        \label{fig:f1}
    }
    \hfil
    \subfloat[mAP]{
        \includegraphics[width=0.48\linewidth]{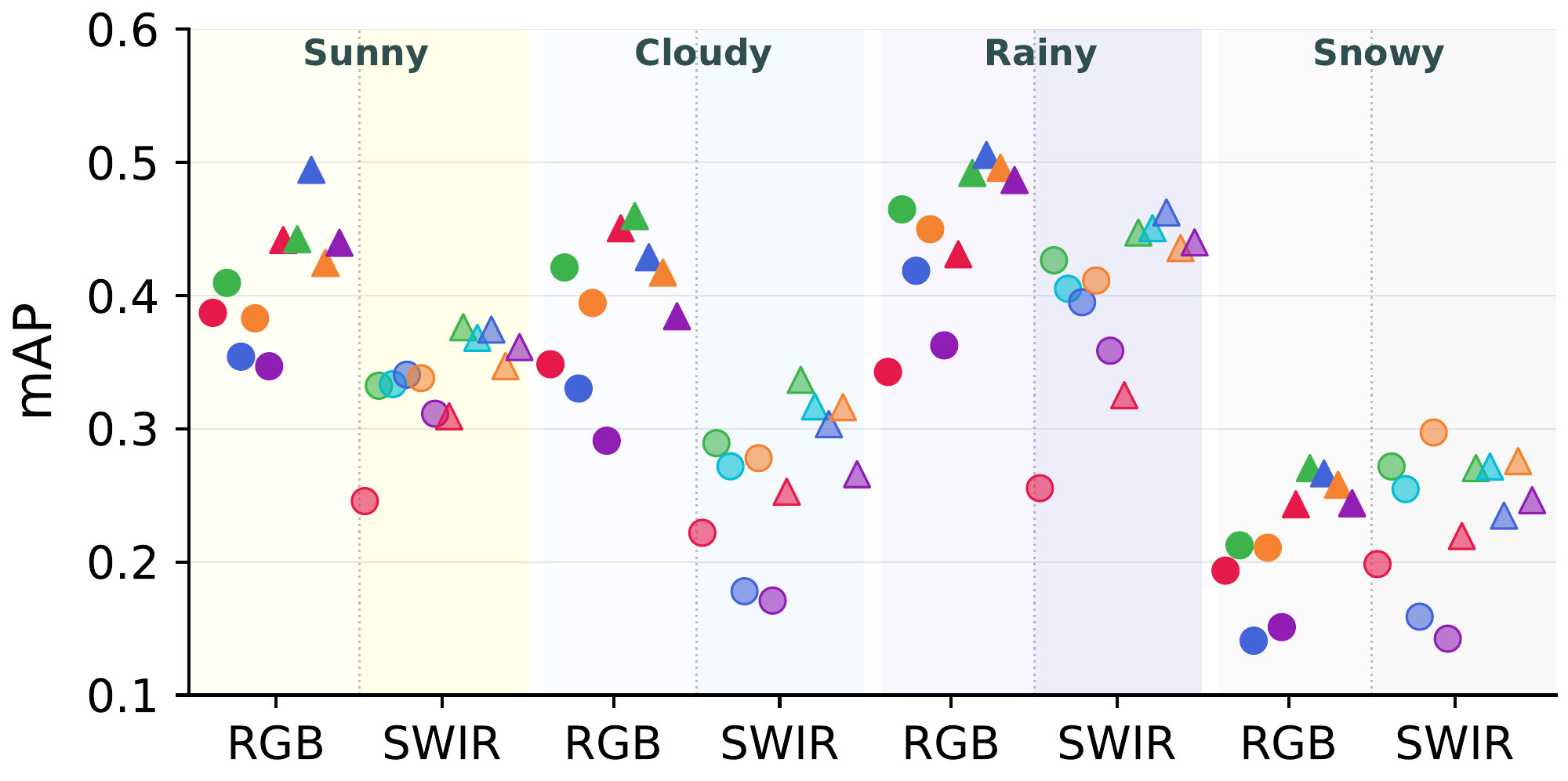}
        \label{fig:map}
    }
    
    \vspace*{-3mm}\includegraphics[width=0.75\linewidth]{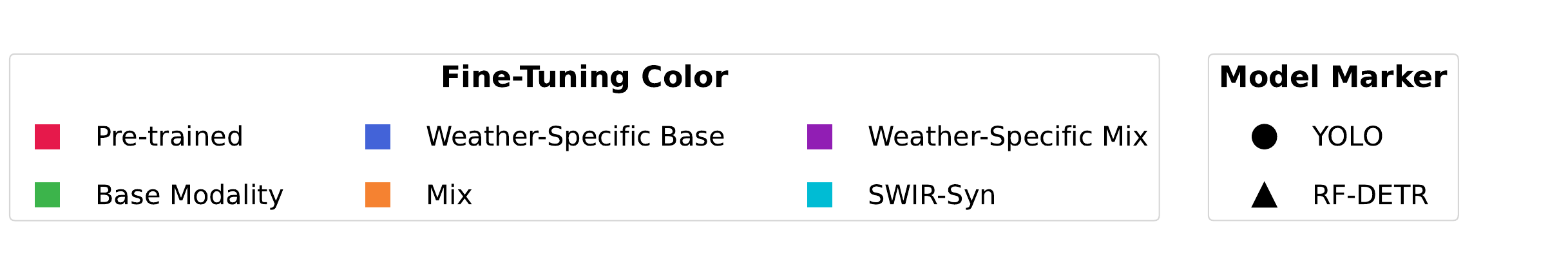}
    \vspace{-5mm}
    \caption{Performance comparison across weather conditions (Sunny, Cloudy, Rainy, Snowy) and sensor modalities (RGB, SWIR) of YOLO and RF-DETR: (a) Weighted Precision, (b) Weighted Recall, (c) Weighted F1-score, and (d) bootstrapped mAP mean. RF-DETR generally outperforms YOLO under the same fine-tuning. Pretrained models show the lower recall than fine-tuned but competitive precision. RGB performs better overall, while RGB and SWIR are comparable under snowy conditions. \textbf{Better visualized in color.}}
    \label{fig:weather_wise_graph_comparison}%
\end{figure*}

\subsection{Combined-Weather Analysis}

\autoref{tab:quantitative_evaluation_merged} presents the results of aggregated weather evaluation on RGB and SWIR images. 
Overall, we observe that RF-DETR consistently outperforms YOLOv8 across both modalities and all fine-tuning strategies. 
For RGB camera, the fine-tuning generally improves \FoneW, \RecallW, and \mAP~over the pre-trained weights for both models. 
For SWIR, these gains from fine-tuning are more significant, due to the large domain gap between visible and SWIR imaging. 
The \PrecW\xspace remains competitive across all configurations in both models and modalities, indicating the gains are largely driven by a reduction in missed detections. 
In all cases, Mix fine-tuning yields competitive performance compared to modality-specific fine-tuning, making it a viable single-model solution for use across both cameras. 
Alongside, SWIR-Syn trained models yield comparable \FoneW, \RecallW, and \mAP~to SWIR-only models with only marginal drops, suggesting synthetic SWIR augmentation as a promising direction to address data scarcity, as also shown in \cite{YannisRFIAP2026}.
Notably, \AP{person} drops after fine-tuning across all configurations and models, suggesting loss of generalization during domain adaptation, possibly due to low support combined with visual distinctiveness of pedestrian compared to other classes. 
However, RGB camera results consistently give higher scores than their SWIR counterparts, indicating that the visible spectrum provides more favorable detection conditions when aggregated over all weather scenarios.

\subsection{Per-Weather Analysis}

\autoref{fig:weather_wise_graph_comparison} presents a per-weather comparison of RGB and SWIR modalities across various fine-tuning strategies. 

Precision (\autoref{fig:precision}) broadly remain competitive across all weather conditions (ranging between 0.73 to 0.93). A relative decline of scores is observed for snowy condition in both the cameras, which is due to complex scenes and reduced contrast during snow. YOLOv8 Weather-Specific Mix fine-tuning is observed to yield low precision with across all four conditions, occasionally accompanied by Weather-Specific Base, whereas RF-DETR maintains competitive scores under same configurations. This suggests YOLO is more susceptible to overfitting under limited per-condition training data. Off-the-shelf pre-trained again gives comparable precision scores to fine-tuned models.

Recall (\autoref{fig:recall}) generally improves with fine-tuning across both models, modalities, and weather conditions, with the exception of RGB images under sunny and cloudy conditions, where fine-tuned models 
yield scores comparable to the pretrained model. This is because the visible spectrum aligns closely with the pretraining domain under good visibility. Under cloudy conditions, RGB models significantly outperform their SWIR counterparts across same fine-tunings, potentially because of clear weather, accompanied by high visibility and strong contrast, whereas this gaps becomes thin for sunny condition due to uneven illumination. Under rainy and snowy conditions, the recall of both sensors remain competitive. Similar to combined evaluation, RF-DETR models are observed to marginally but consistently give better scores than YOLO under same fine-tuning, demonstrating superior choice. Within RF-DETR models, all fine-tunings give comparable recall in all four weather, compared to YOLOv8, where again weather specific mix and occasionally weather specific base degrades in performance.

F-1 score (\autoref{fig:f1}) and bootstrapped mAP mean (\autoref{fig:map}) broadly demonstrate similar trends, and hence are discussed jointly. Overall, training on domain-matched data (RGB trained models on RGB images and SWIR trained models on SWIR camera) gives the good results and maintains stability across most conditions. Weather-specific training again shows divergent behaviors; it results in sub-optimal performance for YOLOv8, whereas it demonstrate high performance fluctuations with RF-DETR across conditions and metrics. With that said, Mix fine-tuning remains consistently close to the top and even achieves the best mAP under snowy condition and SWIR camera using YOLOv8, making it a viable single-model option without significant loss in performance. For modality comparison, RGB again clearly dominates SWIR under cloudy conditions for similar configurations, with the gap progressively narrowing through sunny and rainy conditions. Both sensors become very competitive under snowy conditions, where a clear inference is not possible. This reinforce that harsh weather heavily degrades the visible spectrum. SWIR-Syn models perform at par with SWIR-only trained models across all weather conditions, further supporting synthetic augmentation as a future direction to handle SWIR data scarcity. 

In summary, our combined and weather-wise analysis indicate that fine tuning generally improves detection performance for adverse weather, more significantly for the SWIR domain. All-weather domain-specific fine-tuned models show strong performance, and mix fine-tuning remains a competitive alternative. Conversely, weather-specific fine-tuning exhibits instability. RF-DETR proves more stable than YOLO and generally achieves higher evaluation metrics. However, across both types of anaysis, RGB camera metrics either matches or outperforms the SWIR camera metrics, asserting visible-spectrum imagery as primary and reliable modality for standard road-scene detection. This motivates a dedicated analysis to identify possible scenarios where SWIR camera provides us measurable benefits. The next section focuses on mining such scenarios using mininal human intervention.

%% file: sections/04_sensor_dominance.tex
\section{Beyond Aggregate Metrics: Sensor-Dominance Mining}
\label{sec:mining}

While the previous results establish the general superiority of RGB sensors over SWIR, some rare but safety-critical cases where the SWIR camera could provide us benefits may be masked in global performance averages. In this section, we investigate the complete base dataset to isolate those scenarios. We proposed a framework with three variants based on multi-model consensus to automatically segregate RGB-SWIR image pairs into one of \RGBBetter (RGB detects objects absent in SWIR), \SWIRBetter (SWIR detects objects absent in RGB), \textit{Equal} (both modalities detect same objects), and \textit{Discarded} (no decision can be made) categories. This significantly reduces the search space. Following human curation on the \SWIRBetter set, we categorized the image pairs into four environmental scenarios based on common characteristics. %

\subsection{Automated Segregation Framework}
\label{sec:framework}

We define three model groups, each of which act as an independent voter for each RGB-SWIR image pair: 1) RF-DETR-Mix, which processes both sensor image; 2) YOLO-Mix, which also processes both sensor image; and 3) YOLO-RGB and YOLO-SWIR, where each processes its matching sensor image. 

For voter group $g$ and image pair $p_i$, pseudo-labels are generated for both modalities. Let $\mathcal{D}_g^{\text{RGB}}(p_i)$ and $\mathcal{D}_g^{\text{SWIR}}(p_i)$ denote the respective detection sets. Two detections $d \in \mathcal{D}_g^{\text{RGB}}$ and $d' \in \mathcal{D}_g^{\text{SWIR}}$ are considered a match if they share the same class label and spatially, their bounding boxes satisfy $\mathrm{IoU}(d, d') \geq 0.3$. Let $\mathcal{U}_g^{\text{RGB}}(p_i)$ and $\mathcal{U}_g^{\text{SWIR}}(p_i)$ denote the sets of unmatched detections in each modality after this assignment. The per-group vote 
$\ell_g(p_i) \in \{\mathtt{s},\, \mathtt{r},\, \mathtt{e},\, \mathtt{d}\}$ is then assigned as 
\begin{equation}
    \ell_g(p_i) =
    \begin{cases}
        \mathtt{s} & \text{if } |\mathcal{U}_g^{\text{SWIR}}| > 0 
                     \text{ and } |\mathcal{U}_g^{\text{RGB}}|  = 0 \\[4pt]
        \mathtt{r} & \text{if } |\mathcal{U}_g^{\text{RGB}}|  > 0 
                     \text{ and } |\mathcal{U}_g^{\text{SWIR}}| = 0 \\[4pt]
        \mathtt{e} & \text{if } |\mathcal{U}_g^{\text{SWIR}}| = 0 
                     \text{ and } |\mathcal{U}_g^{\text{RGB}}|  = 0 \\[4pt]
        \mathtt{d} & \text{if } |\mathcal{U}_g^{\text{SWIR}}| > 0 
                     \text{ and } |\mathcal{U}_g^{\text{RGB}}|  > 0 \\
    \end{cases}\text{~.}
    \label{eq:per_voter}
\end{equation}

\noindent
An $\mathtt{s}$ (resp.\ $\mathtt{r}$) vote indicates that group $g$ finds at least one exclusively SWIR (resp.\ RGB) detection with no counterpart in the other modality, while all remaining detections are mutually matched. An $\mathtt{e}$ vote indicates consistent detections between both modality. A $\mathtt{d}$ vote is when both modalities simultaneously produce unmatched detections (e.g., a person detected exclusively in SWIR and a bus detected exclusively in RGB), such that no single modality can be declared dominant. We directly categorize such pairs as \textit{Discarded}, irrespective of the consensus framework applied. For the remaining pairs, let $n_s$, $n_r$, and $n_e$ denote the number of groups returning $\mathtt{s}$, $\mathtt{r}$, and $\mathtt{e}$ votes respectively, where $n_s + n_r + n_e = 3$. Any pair for which $n_s \geq 1$ and $n_r \geq 1$ is simultaneously true is further placed in \textit{Discarded}, as cross-group disagreement on the dominant modality cannot be resolved without a ground truth. The three segregation methods M1, M2, and M3 then apply progressively stricter consensus rules over the remaining pairs to assign each into one of the three remaining categories. For each method M$k \in \{\text{M1}, \text{M2}, \text{M3}\}$, the final category assignment $\mathcal{C}^{\text{M}k}(p_i)$ is thus determined by
\begin{equation}
    \mathcal{C}^{\text{M}k}(p_i) =
    \begin{cases}
        \textit{Discarded} & \text{if } n_s \geq 1 \text{ and } n_r \geq 1 \\[2pt]
        \SWIRBetter & \text{if } n_s \geq k \text{~and~} n_r = 0 \\[2pt]
        \RGBBetter  & \text{if } n_r \geq k \text{~and~} n_s = 0\\[2pt]
        \textit{Equal}         & \text{if } n_e > 3-k   \\
    \end{cases}\text{~.} 
    \label{eq:m1}
\end{equation}
Method M1 (\textit{any vote}) is the most permissive criterion, where a single $\mathtt{s}$ or $\mathtt{r}$ vote sufficient to declare modality dominance. Naturally, only unanimous $\mathtt{e}$ votes result in the \textit{Equal} category.
Method M2 (\textit{majority vote}) applies a majority-vote rule, requiring at least two consistent votes to establish a particular sensor-dominance; otherwise, pairs are categorized as \textit{Equal}.
Method M3 (\textit{unanimous vote}) enforces unanimous agreement, assigning sensor-dominance category only when all three groups agree, and categorizing all other cases as \textit{Equal}.
Overall, M1, M2, and M3 define a spectrum from high recall to high precision, increasing confidence in sensor-dominant cases at the cost of a smaller sensor-dominant set, as low-confidence pairs are reassigned to \textit{Equal}. 
\autoref{tab:methodology_comparison} summarizes the distribution of image pairs obtained by applying M1, M2, and M3 to the dataset.

Across all three methods, the number of \RGBBetter pairs are consistently more than \SWIRBetter pairs, reinforcing the conclusions drawn from the global evaluation that visible-spectrum imaging remains dominant or equal in most scenarios. 
At the same time, a limitation of the proposed framework is that no reliable decision can be made for 61.5\% of the pairs due to simultaneous unmatched detections in both modalities or cross-group disagreement. 
Nevertheless, the methods successfully isolate a measurable subset of \SWIRBetter pairs, suggesting potential advantages for SWIR imaging in specific conditions. 
Therefore, we introduce manual curation to filter out any spurious pairs categorized by algorithmic artifacts, and analyze the true \SWIRBetter cases in greater detail.

\begin{table}[!t]
\centering
\caption{Distribution of number of frame pairs across the three segregation methods. \textbf{\RGBBetter} and \textbf{\SWIRBetter} columns indicate frames where that respective sensor strictly outperformed the other. \textbf{Equal} denotes same performance, while \textbf{Discarded} represents conflicted cases. Percentages represent the proportion of the total dataset.}
\label{tab:methodology_comparison}
\begin{tabular}{@{}lcccl@{}}
 \toprule
 \textbf{Method} & \textbf{\RGBBetter} & \textbf{\SWIRBetter} & \textbf{Equal} & \textbf{Discarded} \\
 \midrule\midrule
 M1: Any Vote & \makecell{14,596 \\ \textit{\scriptsize (14.7\%)}} & \makecell{10,344 \\ \textit{\scriptsize (10.4\%)}} & \makecell{13,358 \\ \textit{\scriptsize (13.4\%)}} & \makecell{61,183 \\ \textit{\scriptsize (61.5\%)}} \\
 \addlinespace
 M2: Majority Vote & \makecell{8,514 \\ \textit{\scriptsize (8.6\%)}} & \makecell{4,945 \\ \textit{\scriptsize (5.0\%)}} & \makecell{24,839 \\ \textit{\scriptsize (25.0\%)}} & \makecell{61,183 \\ \textit{\scriptsize (61.5\%)}} \\
 \addlinespace
 M3: Unanimous Vote & \makecell{3,649 \\ \textit{\scriptsize (3.7\%)}} & \makecell{1,668 \\ \textit{\scriptsize (1.7\%)}} & \makecell{32,981 \\ \textit{\scriptsize (33.2\%)}} & \makecell{61,183 \\ \textit{\scriptsize (61.5\%)}} \\
 \bottomrule
\end{tabular}%
\end{table}

\subsection{Manual Curation Procedure}\label{sec:hils}

While the automated mining framework provides a useful first pass  that identifies modality-dominance trends at scale, all three methods remain susceptible to incorrect modality-dominance assignments due to false positives, arising from model hallucinations and detection errors. To obtain a reliable subset, we focus our analysis on the \SWIRBetter set and apply a manual curation procedure.

Each candidate pair is retained only if it satisfies two criteria. First, the SWIR advantage is verified through manual inspection, ensuring that it contains additional valid detections compared to RGB after removing false positives. Second, to avoid over-representation of near-identical frames from the same scenario, a minimum temporal spacing of 10 frames is enforced between consecutive filtered samples within each sequence.

\begin{table*}[t!]
\centering
\caption{Validation of four SWIR-better filtered scenarios done by RF-DETR (modality-specific models). Best values between modalities are in \textbf{bold}, and only statistically significant classes are considered for class-wise AP. Results indicate a clear SWIR advantage, primarily driven by improved recall.}
\label{tab:rf_detr_specialist_summary}
\footnotesize %
\setlength{\tabcolsep}{5pt} %
\begin{tabular}{@{}llccccccccc@{}} %
\toprule
\multirow{2}{*}{Condition} & \multirow{2}{*}{Modality} & \multicolumn{3}{c}{Weighted Metrics} & \multicolumn{3}{c}{Binary Class Metrics} & \multicolumn{3}{c}{Class-wise AP} \\
\cmidrule(lr){3-5} \cmidrule(lr){6-8} \cmidrule(lr){9-11}
& & Prec. & Recall & F-1 & Prec. & Recall & AP & Car & Truck & Person \\
\midrule\midrule
\multirow{2}{*}{Hidden} 
& RGB & 0.9358 & 0.7631 & 0.8333 & \textbf{0.9454} & 0.7685 & 0.5420 & 0.6095 & 0.5739 & 0.1380 \\
& SWIR & \textbf{0.9370} & \textbf{0.8062} & \textbf{0.8638} & 0.9451 & \textbf{0.8116} & \textbf{0.6212} & \textbf{0.6723} & \textbf{0.6545} & \textbf{0.2978} \\
\midrule
\multirow{2}{*}{Very-Far} 
& RGB & 0.9186 & 0.7288 & 0.8108 & 0.9322 & 0.7408 & 0.5029 & 0.5017 & 0.5088 & - \\
& SWIR & \textbf{0.9296} & \textbf{0.7887} & \textbf{0.8520} & \textbf{0.9536} & \textbf{0.8092} & \textbf{0.5723} & \textbf{0.5753} & \textbf{0.5541} & - \\
\midrule
\multirow{2}{*}{Glare} 
& RGB & 0.8719 & 0.3318 & 0.4746 & 0.8537 & 0.3318 & 0.1950 & 0.1845 & - & - \\
& SWIR & \textbf{0.9203} & \textbf{0.7204} & \textbf{0.8076} & \textbf{0.9212} & \textbf{0.7204} & \textbf{0.3836} & \textbf{0.3631} & - & - \\
\midrule
\multirow{2}{*}{Water-Window} 
& RGB & 0.8897 & 0.6459 & 0.7465 & 0.9058 & 0.6623 & 0.3581 & 0.3682 & - & - \\
& SWIR & \textbf{0.9428} & \textbf{0.8000} & \textbf{0.8641} & \textbf{0.9537} & \textbf{0.8098} & \textbf{0.5455} & \textbf{0.5683} & - & - \\
\bottomrule
\end{tabular}
\end{table*}

\begin{figure*}[t!]
    \centering
    \vspace*{-4mm}\subfloat[Scenario 1: Glare on the windshield.]{
        \includegraphics[width=0.185\textwidth]{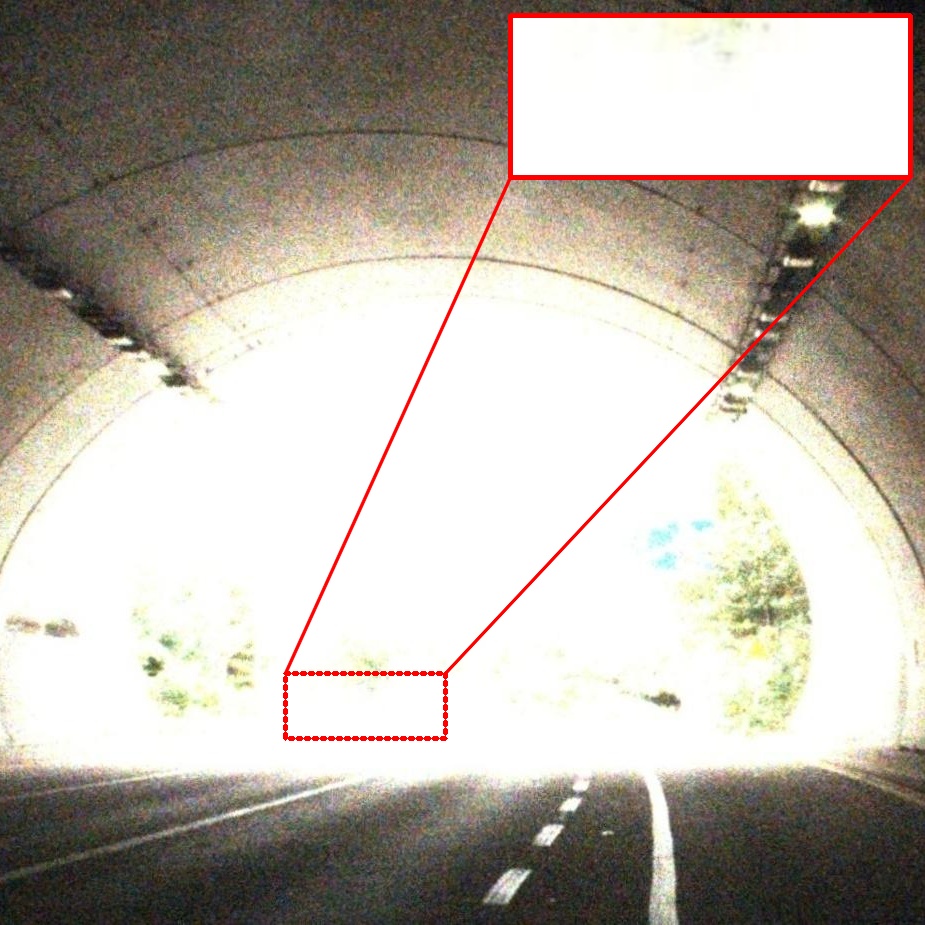}\hfill
        \includegraphics[width=0.185\textwidth]{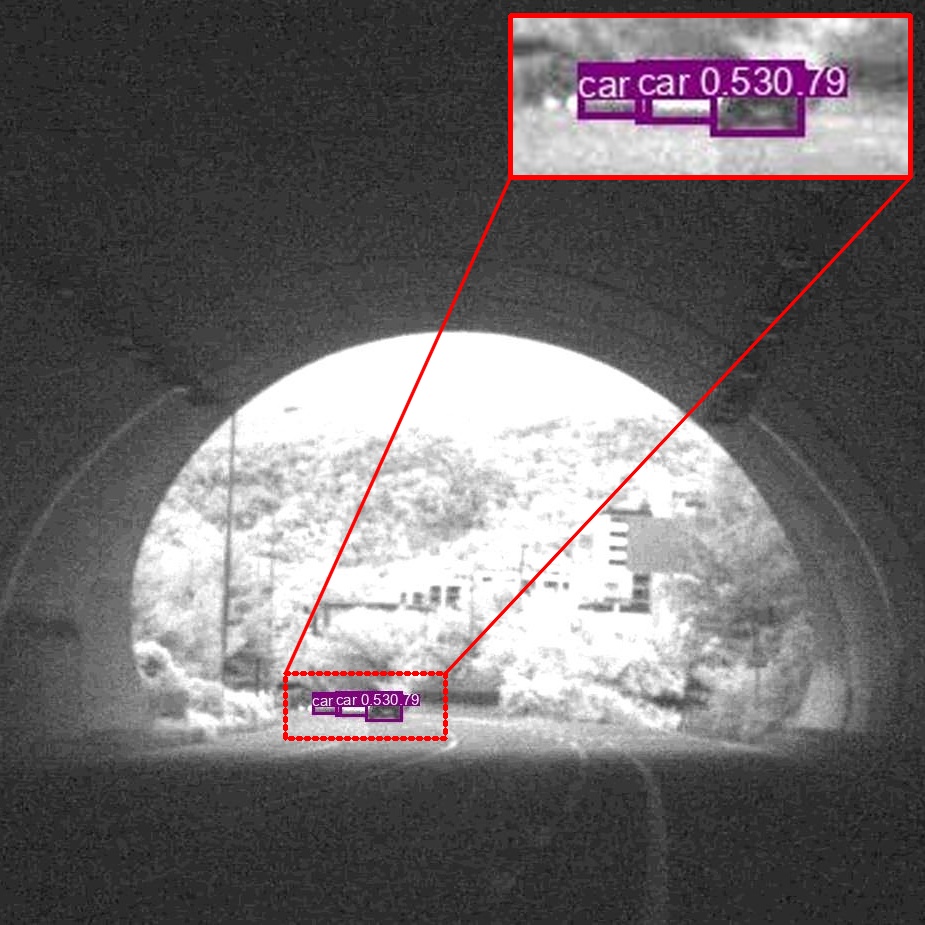}
        \label{fig:scenario_glare}
    }
    \hspace{0.2cm} %
    \subfloat[Scenario 2: Water on the windshield.]{
        \includegraphics[width=0.185\textwidth]{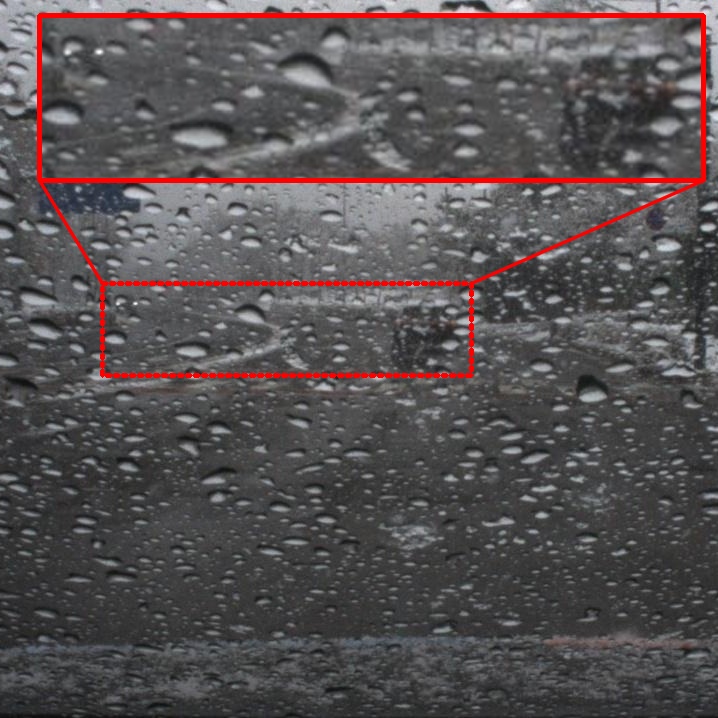}\hfill
        \includegraphics[width=0.185\textwidth]{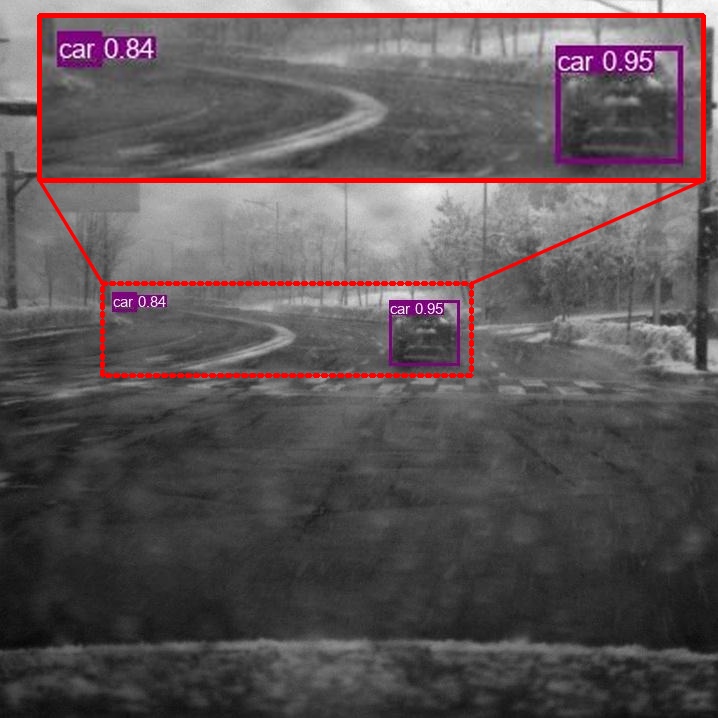}
        \label{fig:scenario_water}
    }
    \\ 
    \vspace*{-0.1cm} 
    
    \subfloat[Scenario 3: Hidden due to contrast.]{
        \includegraphics[width=0.185\textwidth]{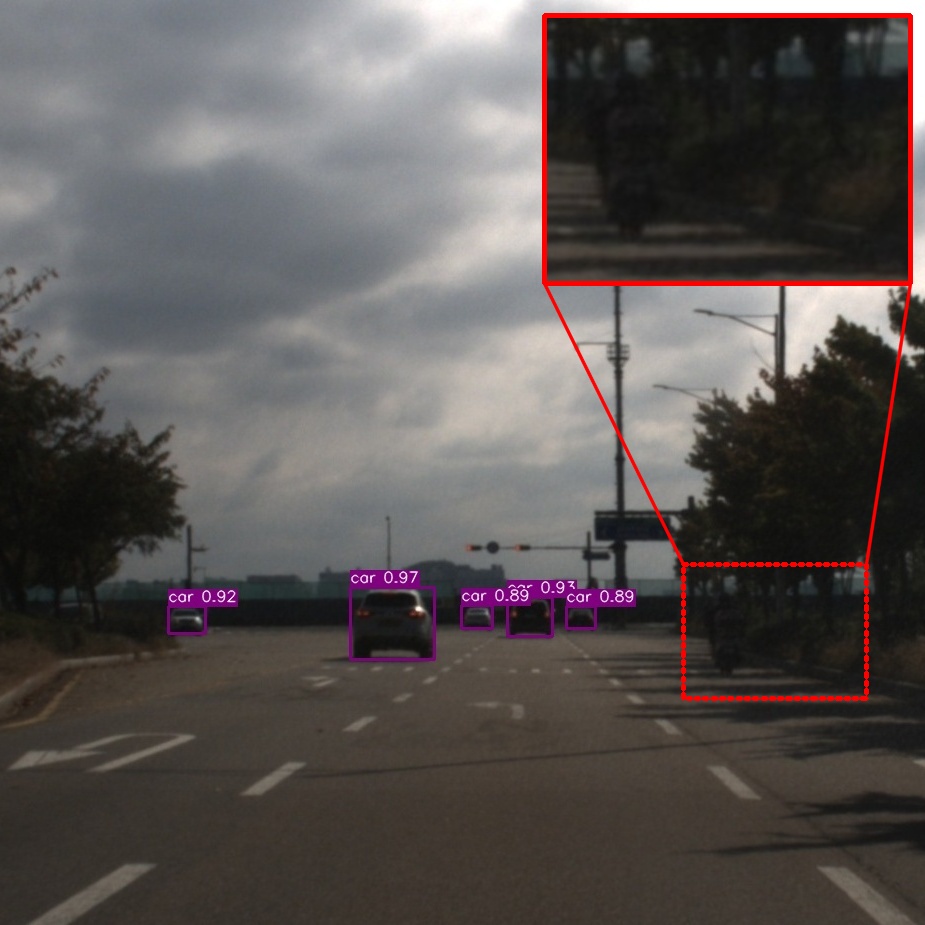}\hfill
        \includegraphics[width=0.185\textwidth]{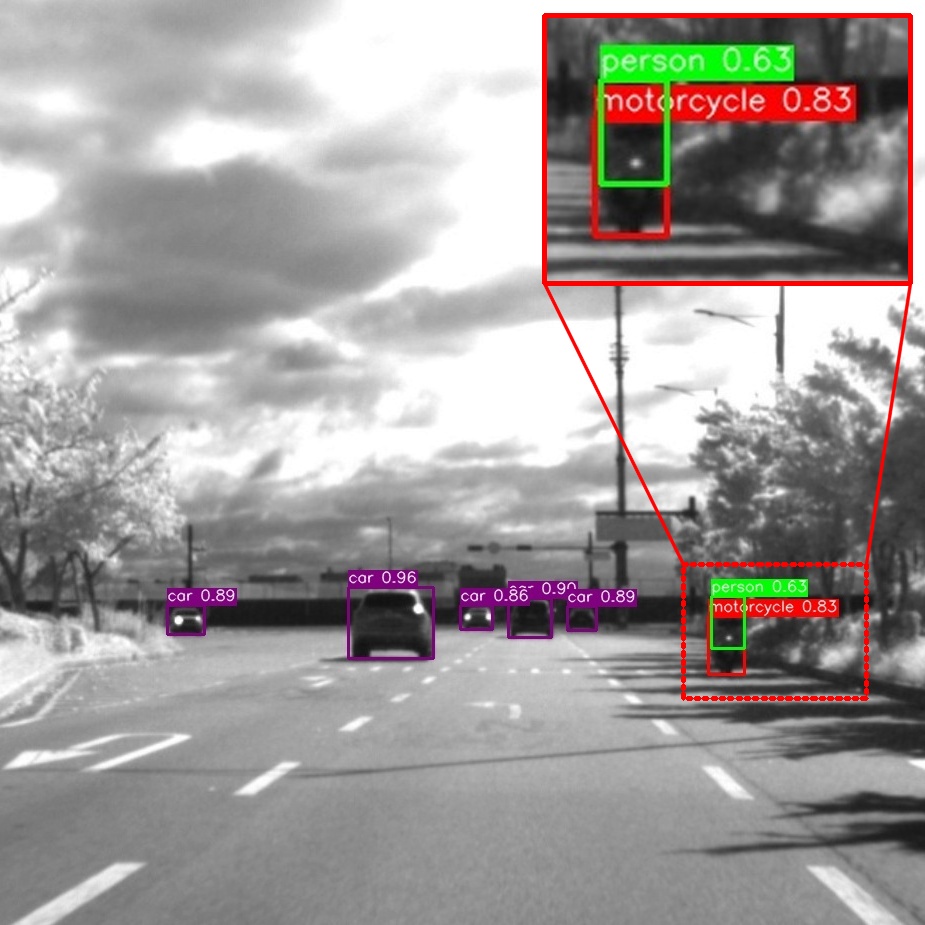}
        \label{fig:scenario_contrast}
    }
    \hspace{0.2cm}
    \subfloat[Scenario 4: Distant Vehicle.]{
        \includegraphics[width=0.185\textwidth]{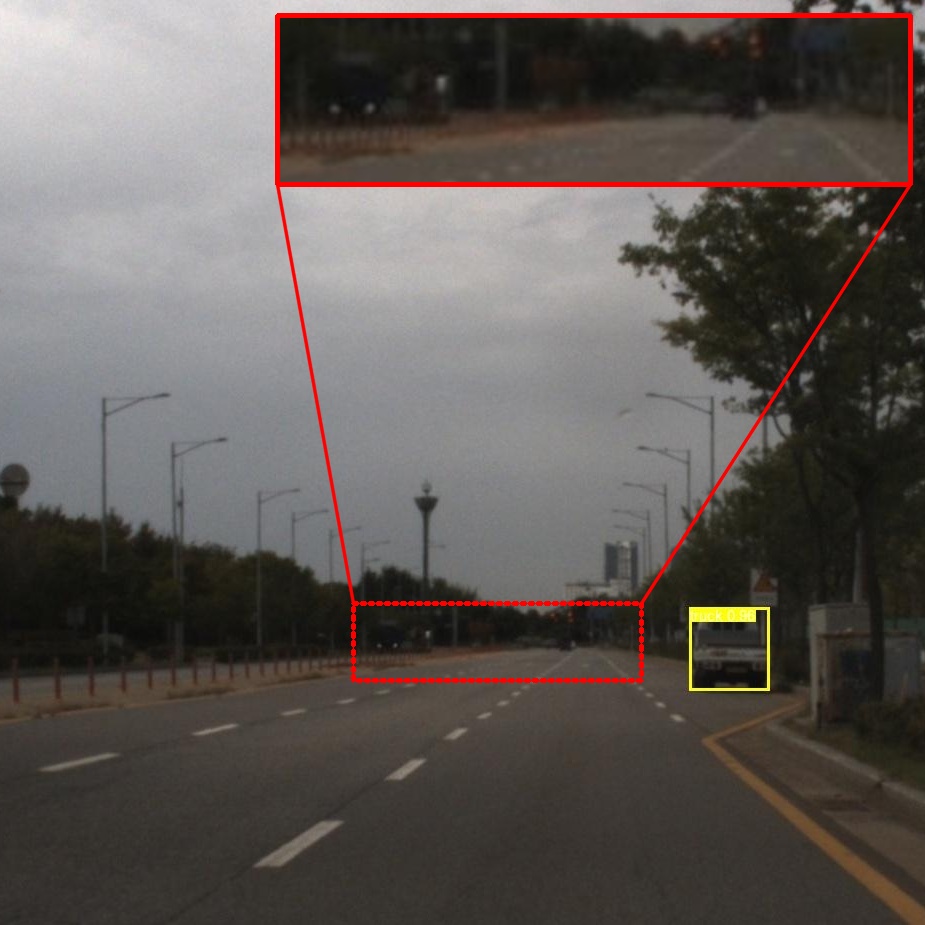}\hfill
        \includegraphics[width=0.185\textwidth]{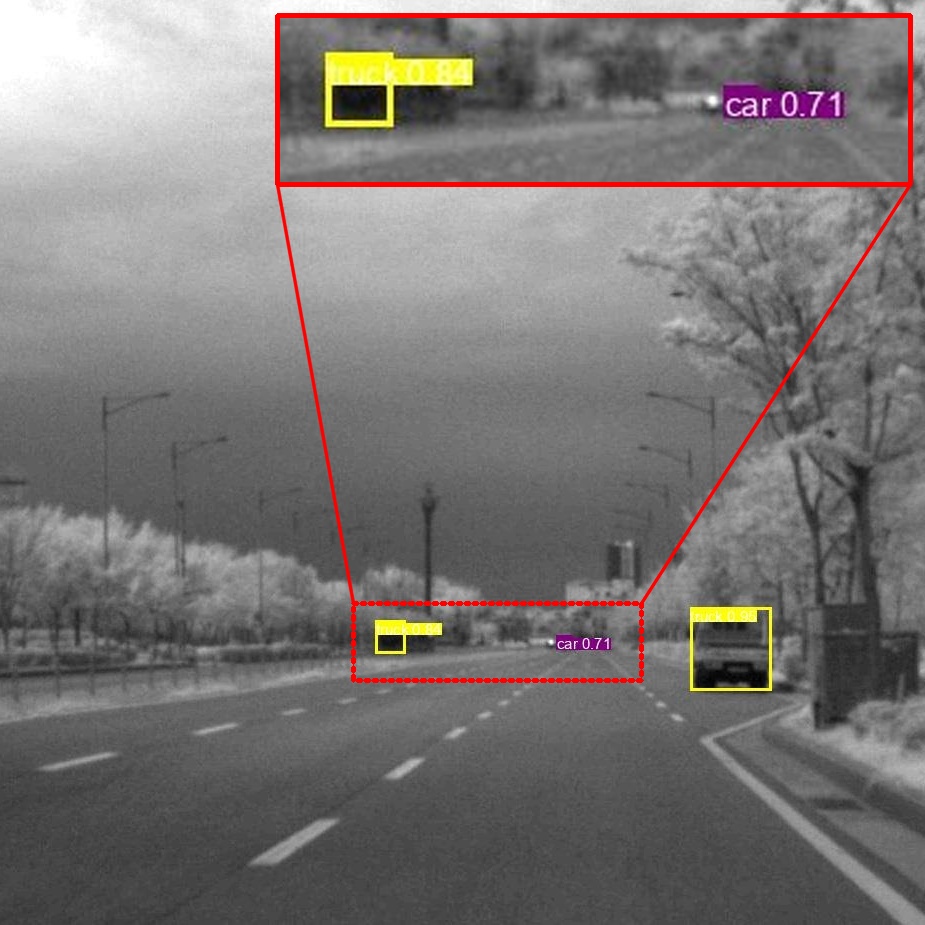}
        \label{fig:scenario_far}
    }
    \vspace{0.1cm} 
    
    \caption{Visual comparison of detections in four SWIR-advantageous mined scenarios produced by RF-DETR-Mix model: (a) glare on the windshield, where glare saturates RGB while SWIR enables vehicle detection; (b) water on the windshield, preventing detections in RGB due to scattering while vehicles are detected in SWIR; (c) motorcyclist hidden due to low contrast in RGB but detected in SWIR, and (d) distant vehicles, where reduced atmospheric scattering in SWIR enables detection of a car and a truck. Each pair shows RGB (left) and SWIR (right). Images are cropped and zoomed for visualization clarity.}
    \label{fig:scenario_bucket_images}
\end{figure*}

Following this process, we obtain a curated set of 532 distinct scene image pairs, {that we named RASMD-HiLS (for \textit{Human in the Loop Segregation})}.
Upon visual inspection, these samples exhibit four recurring environmental scenarios: \textbf{(1)} glare on the windshield, \textbf{(2)} water droplets on the windshield, \textbf{(3)} object hidden due to contrast, and \textbf{(4)} distant vehicles. 
Finally, all selected frames are also manually annotated following a common ground truth for both modalities, as described in \autoref{subsec:dataset}.
This results in a high-quality SWIR-beneficial dataset, which is used for subsequent analysis and validation.

\subsection{Scenario-wise Analysis of SWIR-Advantage Cases}\label{sec:swir-plus-valid}

Following the construction of the curated SWIR-superior RASMD-HiLS dataset, the samples are grouped into the four identified environmental scenarios, and each of them is analyzed independently. Note that these scenarios are not mutually exclusive, and a single image may belong to multiple scenario buckets. To validate the reliability of these scenario-specific subsets, we employ RF-DETR-RGB and RF-DETR-SWIR on respective modality images, both of which were held-out in the filtering process. 
\autoref{tab:rf_detr_specialist_summary} provides quantitative validation using weighted, binary, and class-wise metric of the same. For the class-wise \APc, only classes with significant supports have been taken into consideration. The results confirm that the mined dataset captures common scenario-dependent SWIR modality advantages. 
\autoref{fig:scenario_bucket_images} presents visual comparisons across the four scenarios.
We will now discuss each scenario in detail.

\noindent \subsubsection{Glare on the windshield} 

This scenario corresponds to extreme visible-light saturation, typically caused by direct sunlight reflections when transitioning from low to high illumination conditions (e.g., just before exiting a tunnel). As shown visually in \autoref{fig:scenario_glare}, such saturation significantly degrades RGB image quality, leading to missed detections, while SWIR remains largely unaffected. Quantitatively, \autoref{tab:rf_detr_specialist_summary} validates this across all evaluation metrics in this setting, with substantial gains in recall(s) and AP(s). Notably, this scenario exhibits the largest performance gap between the two modalities.

\begin{figure*}[t!]
    \centering
    
    \begin{tabular}{@{} c @{\hspace{3pt}} c @{\hspace{-29pt}} c @{}}
        & RF-DETR & YOLOv8x \\[-1.0ex] %
        
        \rotatebox[origin=c]{90}{Mix} &
        \raisebox{-0.5\height}{
            \includegraphics[width=0.45\textwidth]{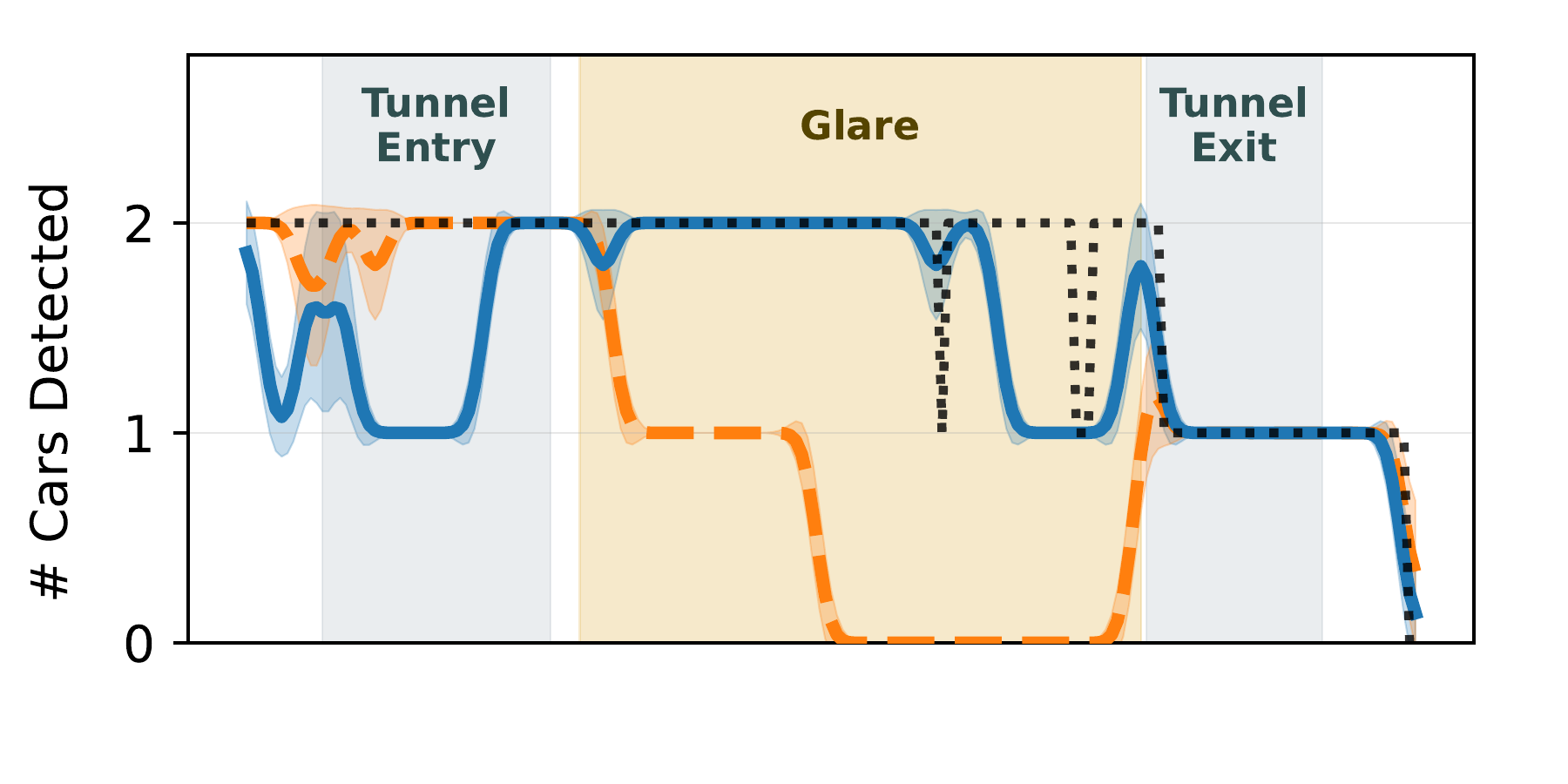}
            \label{fig:graph_rf_mix}
        } &
        \raisebox{-0.5\height}{
            \includegraphics[width=0.45\textwidth]{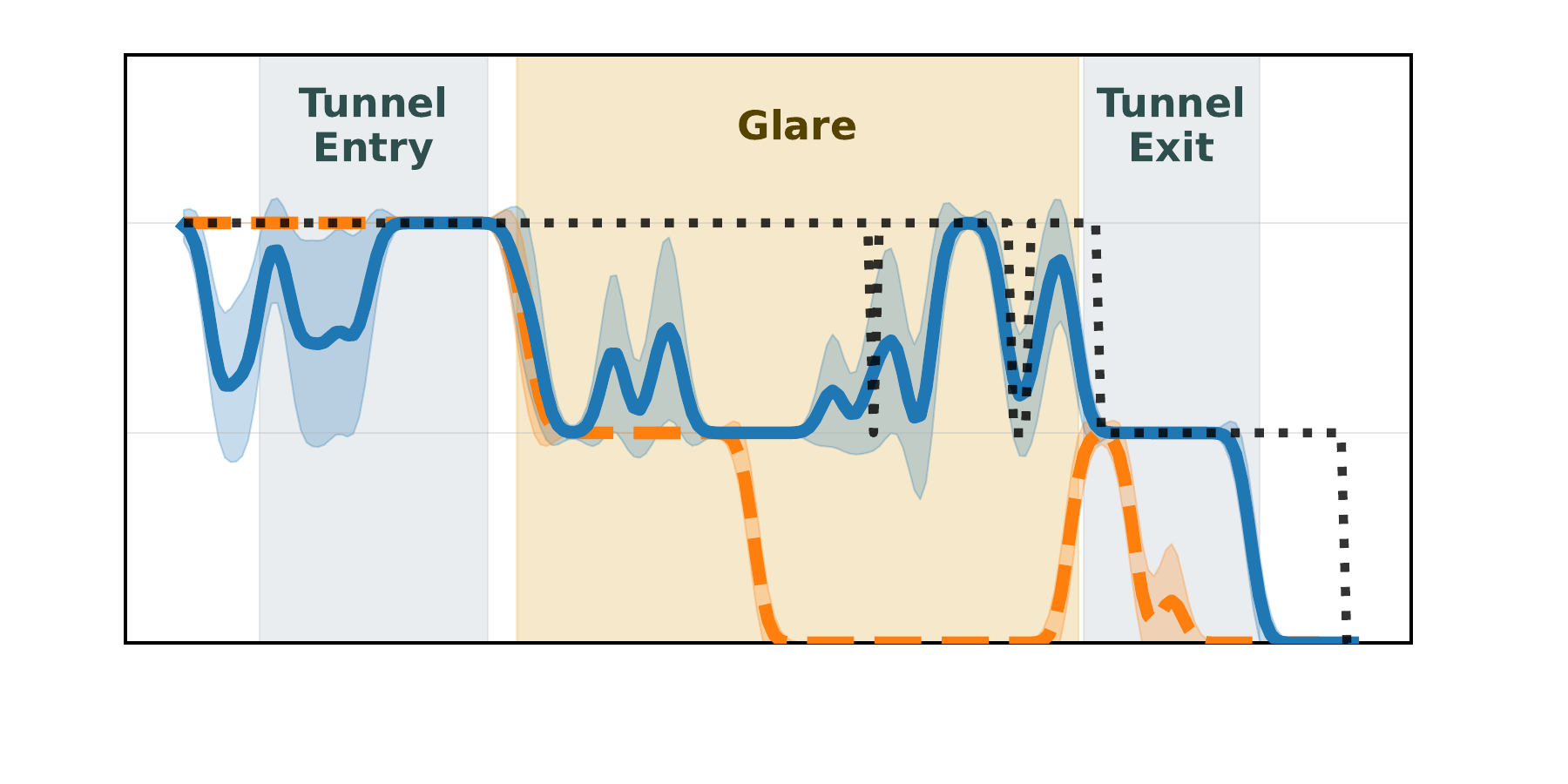}
            \label{fig:graph_yolo_mix}
        } \\[-3.0ex]
        
        \rotatebox[origin=c]{90}{Modality-Specific} &
        \raisebox{-0.5\height}{
            \includegraphics[width=0.45\textwidth]{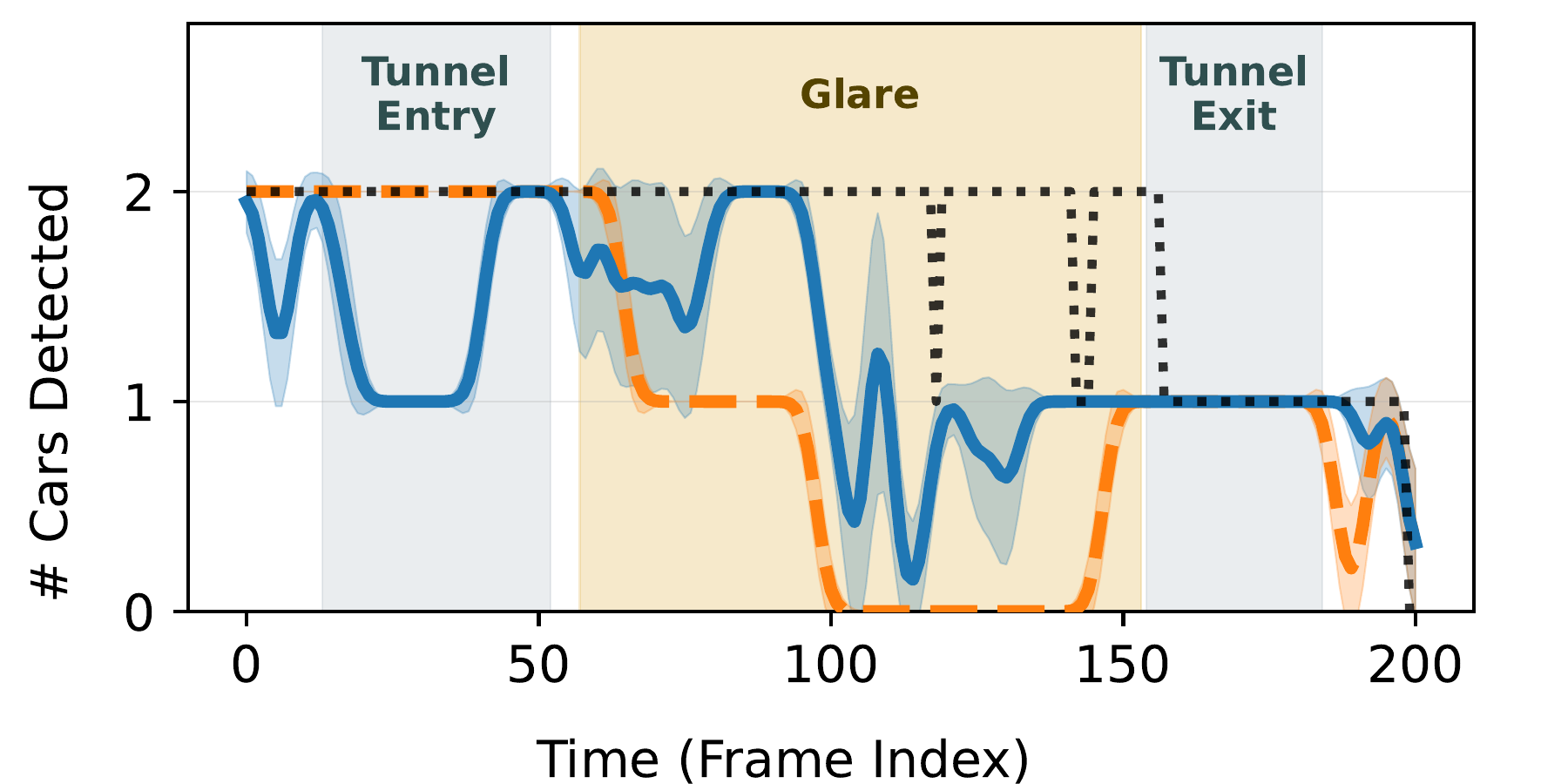}
            \label{fig:graph_rf_spec}
        } &
        \raisebox{-0.5\height}{
            \includegraphics[width=0.45\textwidth]{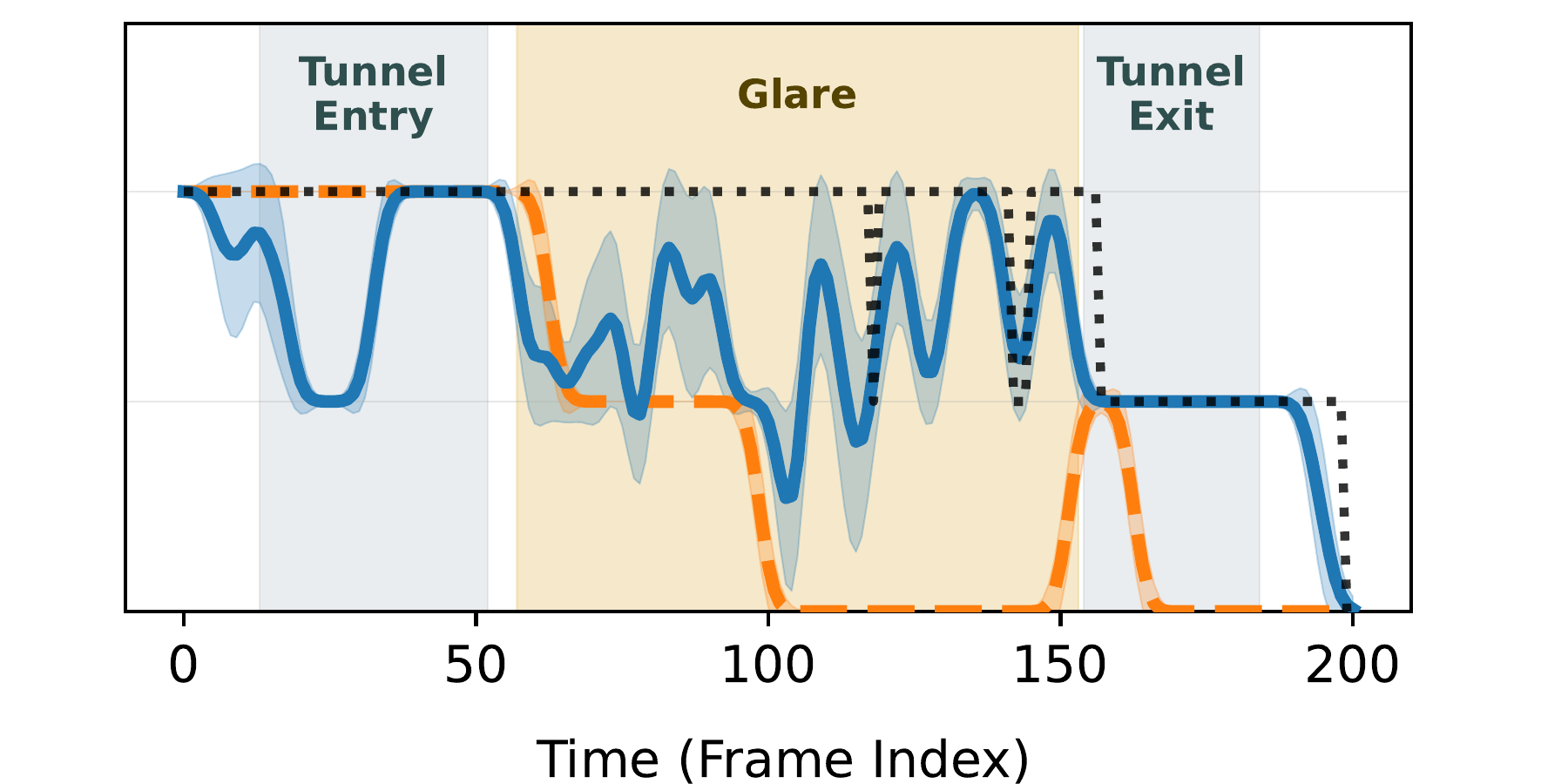}
            \label{fig:graph_yolo_spec}
        } \\
    \end{tabular}
    
    \vspace{0.2cm}
    \includegraphics[width=0.6\textwidth]{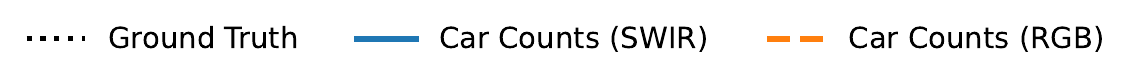} 
    \vspace{-0.1cm} 
    
    \subfloat[Tunnel Entry (Left: RGB, Right: SWIR)]{
        \includegraphics[width=0.150\textwidth]{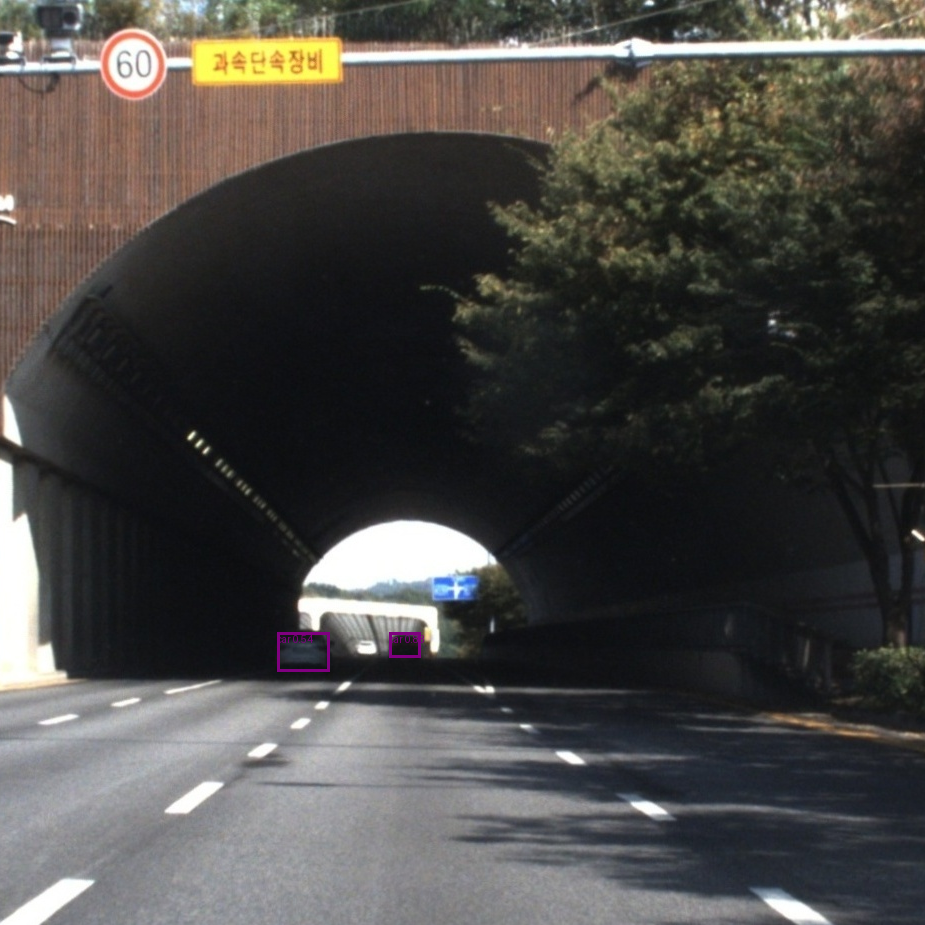}\hfill
        \includegraphics[width=0.150\textwidth]{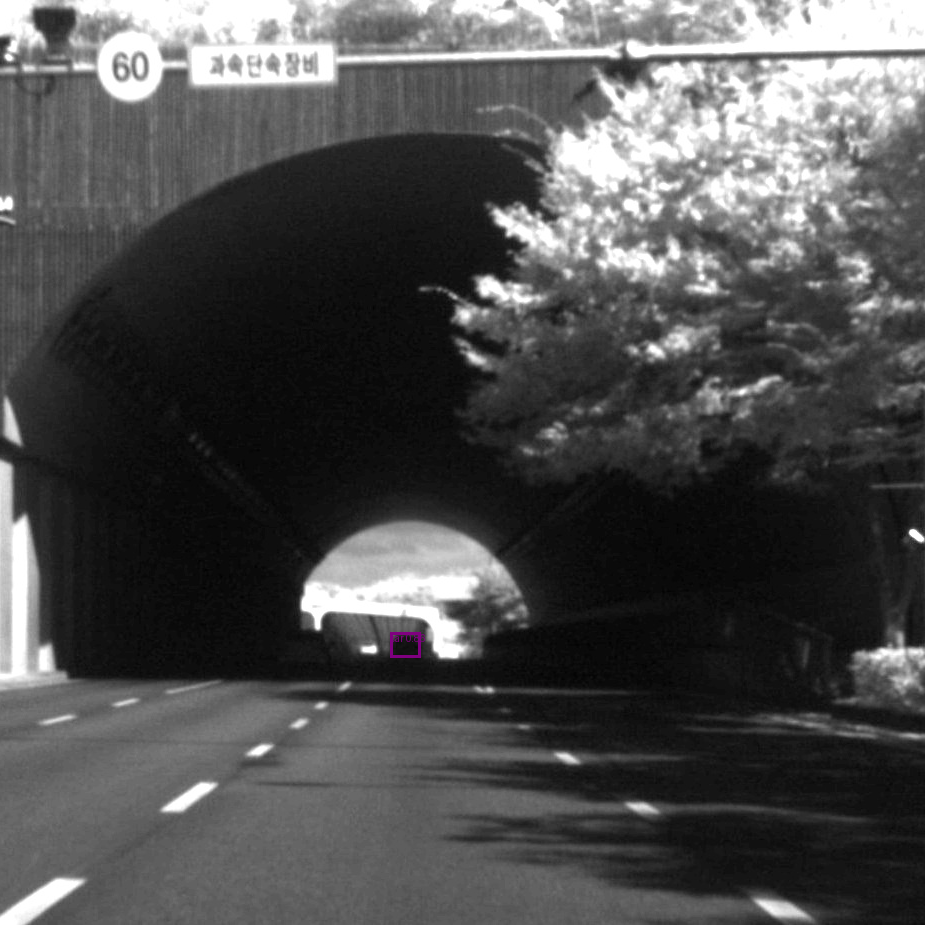}
        \label{fig:div_entry}
    }
    \hfil
    \subfloat[Glare (Left: RGB, Right: SWIR)]{
        \includegraphics[width=0.150\textwidth]{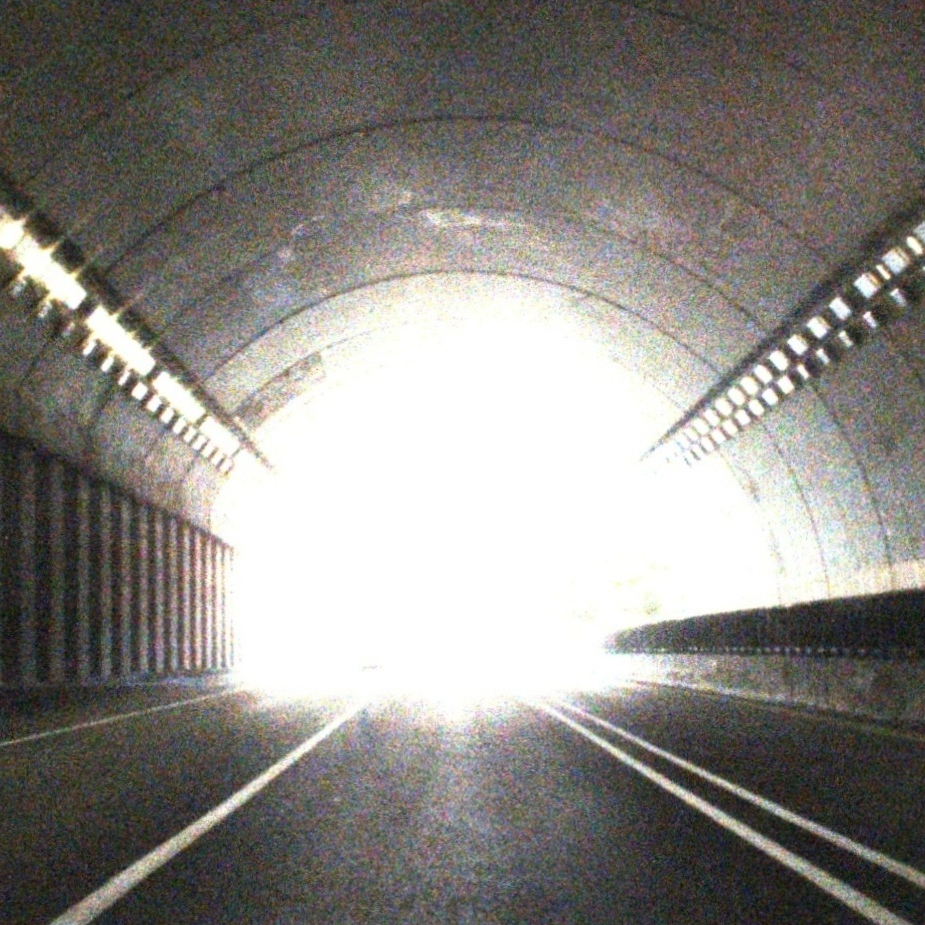}\hfill
        \includegraphics[width=0.150\textwidth]{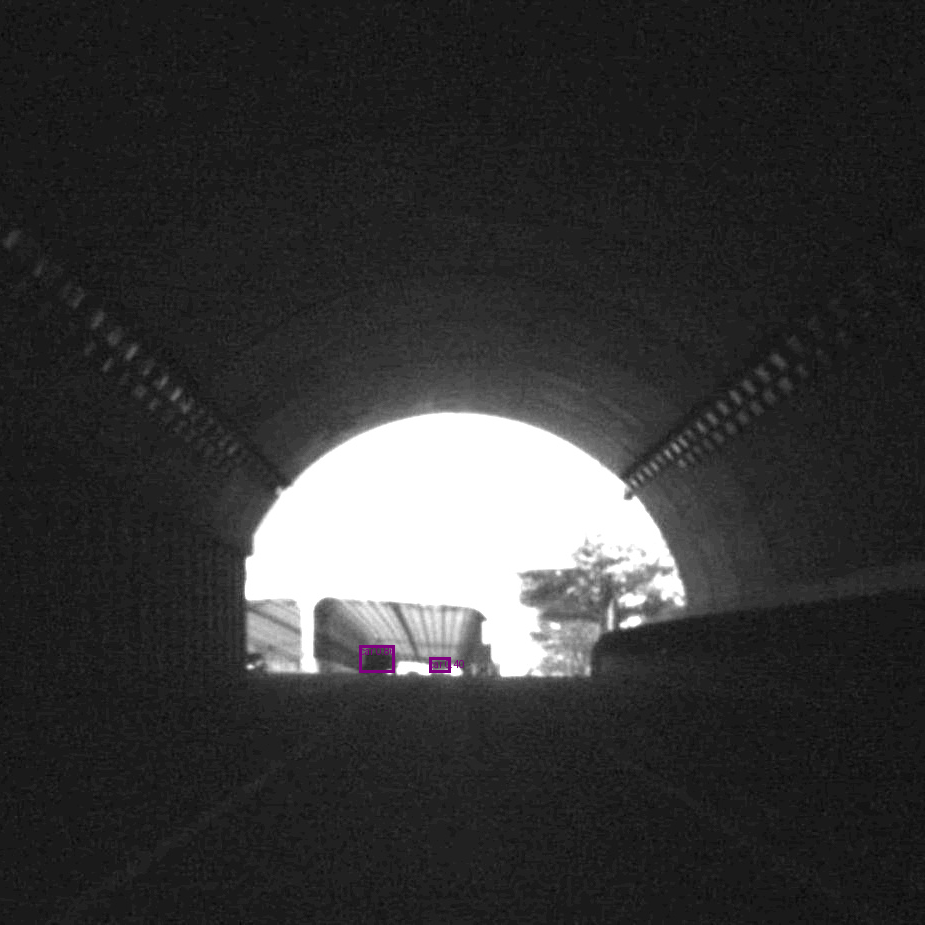}
        \label{fig:div_glare}
    }
    \hfil
    \subfloat[Tunnel Exit (Left: RGB, Right: SWIR)]{
        \includegraphics[width=0.150\textwidth]{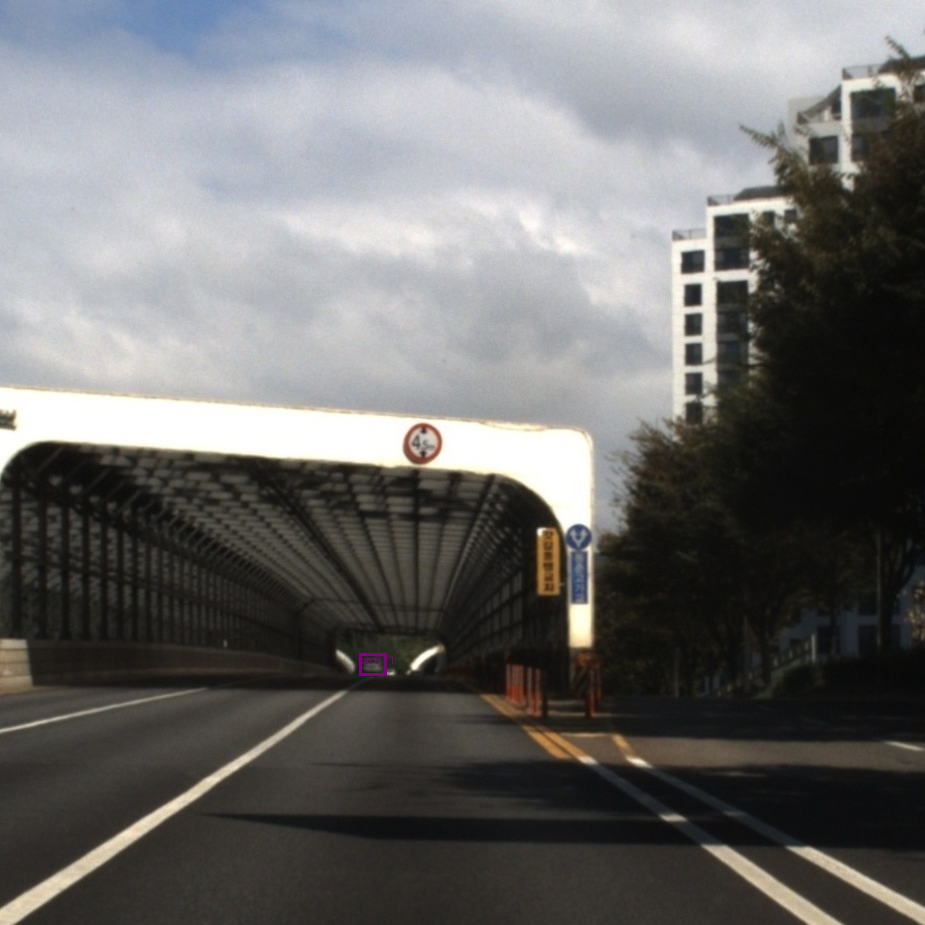}\hfill
        \includegraphics[width=0.150\textwidth]{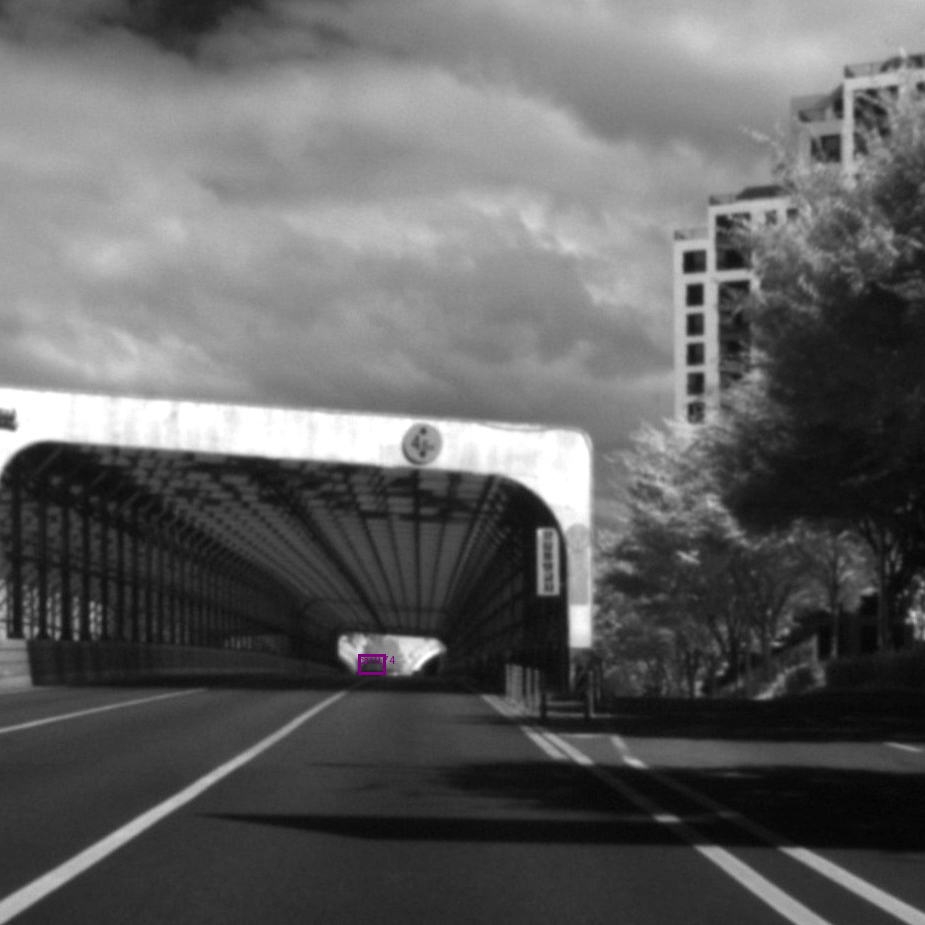}
        \label{fig:div_exit}
    }

    \caption{Temporal analysis of vehicle detection during tunnel traversal, highlighting SWIR robustness to glare. Graphs above shows detection counts from RGB and SWIR for two vehicles using RF-DETR and YOLOv8x under Mixed and modality-specific fine-tuning (curves smoothed with $\sigma = 2$ for visualization). (a) Tunnel entry: due to the rapid transition from bright to dark regions, RGB detects two vehicles while SWIR detects only one; (b) Glare region: RGB pixels saturates while SWIR detects both vehicles; (c) Tunnel exit: under normal illumination, both sensors detect one vehicle.}
    \label{fig:tunnel_evaluation}
\end{figure*}

To further analyze this behavior beyond the curated dataset, we perform a temporal study on a sequence of approximately 200 frames of a vehicle entering and exiting a tunnel. We track detections counts of two leading vehicles over time across both sensors. 
\autoref{fig:tunnel_evaluation} illustrates the results for the four model groups (mixed and modality-specific versions of both RF-DETR and YOLOv8). 
In all graphs, the modality advantage trend varies consistently across the tunnel transition, with RF-DETR-Mix exhibiting the most stable performance. During tunnel entry (\autoref{fig:div_entry}), RGB briefly outperforms SWIR, as the scene transitions from high to low illumination and one of the vehicles lies in a darker region, where it remains distinguishable in RGB due to stronger appearance cues (color and texture), but is less discernible in the grayscale SWIR image. However, once the vehicle reaches near the other end of the tunnel, glare hits (\autoref{fig:div_glare}), and RGB performance drops sharply due to saturation effects, whereas SWIR remains comparably stable. At the tunnel exit under normal illumination (\autoref{fig:div_exit}), the performance of both modalities becomes comparable. The ground truth has again been marked using a unified human view across both modalities.

\subsubsection{Water on the windshield}

This scenario encompasses adverse conditions such as rain or snow, where water droplets accumulate on the windshield, introducing scattering, refraction, and localized blur that significantly degrade visibility in the RGB domain. As illustrated in \autoref{fig:scenario_water}, this often leads to missed detections. In contrast, SWIR is less affected by droplet-induced scattering and can better penetrate such obstructions, preserving scene structure and enabling more reliable detections, consistent with findings in \cite{YEO2026119654}. This behavior is further validated in \autoref{tab:rf_detr_specialist_summary}, where this scenario exhibits the second-largest performance gap after glare, with notable improvements in recall(s) and AP(s) for SWIR. 

\subsubsection{Object hidden due to contrast}

Unlike visibility degradation caused purely by environmental obstructions, this scenario arises from intrinsic scene properties where objects have low contrast against their background. In such cases, similar color distributions, shadows, or weak texture in RGB cause objects to blend into the scene, leading to missed detections, as demostrated for a motorcyclist in \autoref{fig:scenario_contrast}. SWIR, on the other hand, leverages material reflectance differences, which improves the foreground–background separation making such low contrast objects easier to detect. Quantitatively, both modalities show comparable precision, while SWIR reduces missed detections. The improvement is particularly significant for the person class, highlighting pedestrians are impacted by such low-contrast conditions.

\subsubsection{Vehicle at a distance}

This scenario captures long-range detection, where small object size and atmospheric effects such as haze reduce visibility in RGB. While RGB can sometimes also retain advantages due to color and appearance properties, its performance degrades more rapidly under such conditions. As shown in \autoref{fig:scenario_far}, SWIR being less sensitive to visible-spectrum scattering and color attenuation, provides more stable signal characteristics at longer ranges, enabling improved detection of distant vehicles. This is further supported by \autoref{tab:rf_detr_specialist_summary}, where SWIR demonstrates improved detection performance under degraded visibility, although it may not be universally superior in all long-range scenarios.

Overall, while less frequent in real-world driving scenarios, these four scenarios are safety-critical, and SWIR offers a meaningful detection advantage precisely where RGB fails.

%% file: sections/05_discussion.tex
\section{Discussions and Limitations of the study}
\label{sec:discussion}

\noindent\textbf{Comparison of detection performances.} 
Across all fine-tuning strategies and weather conditions (\autoref{tab:quantitative_evaluation_merged}, \autoref{fig:weather_wise_graph_comparison}), RGB generally matches or outperforms SWIR, likely due to richer appearance cues and higher channel diversity. 
In contrast, SWIR benefits more strongly from fine-tuning, reflecting a larger domain gap from visible-domain pretraining. 
Architecturally, the transformer-based RF-DETR consistently outperforms the CNN-based YOLOv8 across modalities, weather conditions, and training strategies, establishing it as the more robust framework. 
Mixed fine-tuning with equal proportions of RGB and SWIR emerges as an effective single-model solution, encouraging modality-invariant representations and achieving competitive cross-modal performance, a trend further confirmed by the temporal tunnel evaluation (\autoref{fig:tunnel_evaluation}). 
Synthetic SWIR augmentation also proves promising in alleviating data scarcity, whereas weather-specific fine-tuning, particularly for YOLOv8, produces inconsistent results, likely due to overfitting and limited generalization.\\

\noindent\textbf{Identification of SWIR-advantage cases.} 
Although 61.5\% of pairs are labeled as \textit{Discarded}, the sensor-dominance mining framework (\autoref{sec:framework}) successfully isolates SWIR-advantageous cases across its three variants (\autoref{eq:m1}), offering a trade-off between selection accuracy and review effort. 
Manual inspection reveals four recurrent scenarios (\autoref{fig:scenario_bucket_images}) where SWIR improves recall while maintaining comparable precision.
The most pronounced gains occur under glare, windshield water droplets, low-contrast pedestrian scenes, and long-range vehicle detection, where RGB performance degrades due to saturation, scattering, low visual contrast, or atmospheric effects, while SWIR remains comparatively robust.\\

\noindent\textbf{Limitations.} This study has several limitations that should be considered when interpreting the results. First, the dataset which reflects real-world driving conditions exhibits strong class imbalance. This, combined with the presence of sequential frames, makes it challenging to ensure sufficient scene diversity and class representation during training, and can lead to overfitting for some underrepresented classes (e.g., \text{AP}\textsubscript{person} in \autoref{tab:quantitative_evaluation_merged}). Second, the proposed sensor-dominance mining framework operates without ground truth and relies on multi model agreement, which may introduce bias toward model-specific behaviors. As a result, a significant portion of data is discarded during filtering, potentially overlooking scenarios even rare in number where SWIR could provide additional benefits. Finally, the mined \SWIRBetter subsets inherit the class distribution of the underlying dataset, where certain conditions (e.g., glare) are dominated by only certain classes (e.g., cars). This necessitates us to use weighted or binary metrics and validating only the statistically significant classes.

%% file: sections/06_conclusion.tex
\section{Conclusion}
\label{sec:conclusion}

The results of this study provide a comprehensive evaluation of RGB and SWIR sensing modalities for object detection under real-world driving conditions, and surface several findings for practical implications of both sensors. 
While aggregate performance indicates the dominance of RGB as the primary modality for general road-scene detection, such global metrics fail to capture rare but safety-critical conditions where SWIR could be useful. 
Based on speed, accuracy, and resources, RF-DETR with mixed fine-tuning emerges as a well-balanced cross modal choice. 
The proposed sensor-dominance mining framework, combining multi-model consensus with targeted manual curation, serves as a scalable approach to segregate modality-specific advantageous pairs from large paired unannotated datasets, and its effectiveness can be further extended with the availability of higher-quality training data and more advanced detection models. 
Within the scope of this study, the framework isolates four SWIR-advantageous conditions: windshield glare, water droplets on the windshield, object visibility under low contrast, and distant vehicle perception, each of which is further validated through quantitative analysis. 
Among these, glare represents the most severe failure of RGB while low-contrast conditions majorly impact pedestrian detection, making SWIR particularly valuable for vulnerable road user safety.

These strengths motivate future work on adaptive fusion strategies that dynamically leverage SWIR under challenging environmental conditions \cite{riffard2026enhancing}. Improving synthetic-to-real alignment for SWIR image generation also remains an open challenge, with future directions including the development of SWIR-specific generative models and broader validation across more diverse datasets and sensor configurations.

\section*{Acknowledgments}

This work benefited from the support of the International Center for Research ``Innovative Systems for Transport and
Production'' of I-SITE CAP 20-25. The authors would also like to thank Trong Nghia Ngo and  Gautier Claisse for their insightful discussions.